\documentclass[lettersize,journal]{IEEEtran}

\usepackage{mathtools}      
\usepackage{amsthm}
\usepackage{amsfonts}
\newtheorem{definition}{Definition}
\usepackage{xspace}
\newcommand{\name}{DyMETER\xspace}

\usepackage{colortbl}
\usepackage{xcolor} 

\usepackage{amsopn}

\usepackage[linesnumbered,ruled,vlined]{algorithm2e}

\usepackage{makecell}
\usepackage{color}
\newcommand{\highlight}[1]{\textbf{#1}}

\newcommand{\blue}[1]{{\color{black}#1}}
\newcommand{\revise}[1]{{\color{black}#1}}
\newcommand{\pamirevise}[1]{{\color{black}#1}}
\newcommand{\pamirevises}[1]{{\color{black}#1}}
\newcommand{\violet}[1]{{\color{black}#1}}
\newcommand{\ignore}[1]{}

\newcommand{\powerpoint}[1]{\textbf{#1}}

\usepackage{enumitem}
\newcommand{\squishlist}
{
	\begin{list}{$\bullet$}
		{
			\setlength{\itemsep}{0pt}
			\setlength{\parsep}{3pt}
			\setlength{\topsep}{3pt}
			\setlength{\partopsep}{0pt}
			\setlength{\leftmargin}{1.5em}
			\setlength{\labelwidth}{1em}
			\setlength{\labelsep}{0.5em}
		}
	}
	
	\newcommand{\squishend}
	{
	\end{list}
}

\usepackage{multirow}
\usepackage{arydshln}
\usepackage{caption}
\usepackage{subcaption}
\usepackage{makecell}           
\usepackage{booktabs}  
\newsavebox{\measurebox}

\usepackage{multicol}

\newcommand{\unsim}{\mathord{\sim}}

\usepackage{url}

\usepackage[ruled,vlined,linesnumbered]{algorithm2e} 
\DontPrintSemicolon 
\SetKwInOut{Input}{Input}
\SetKwInOut{Output}{Output}
\SetKwInOut{Initialization}{Initialization}

\usepackage{dsfont}

\usepackage{balance}

\usepackage{amsmath}

\DeclareMathOperator*{\argmin}{arg\,min}

\begin{document}

\title{Catching Every Ripple: Enhanced Anomaly Awareness via Dynamic Concept Adaptation}

\author{Jiaqi Zhu, Shaofeng Cai, Jie Chen,~\IEEEmembership{Fellow,~IEEE}, Fang Deng,~\IEEEmembership{Fellow,~IEEE}, \\
Beng Chin Ooi,~\IEEEmembership{Fellow,~IEEE}, and Wenqiao Zhang
\thanks{
This work was supported in part by the National Natural Science Foundation of China National Science Fund for Young Scientists (Ph.D.) under Grant 624B2027,
in part by the National Science and Technology Major Project under Grant 2022ZD0119701,
in part by the National Natural Science Foundation of China National Science Fund for Distinguished Young Scholars under Grant 62025301.} 
\thanks{Jiaqi Zhu and Fang Deng are with the School of Automation and the National Key Laboratory of Autonomous Intelligent Unmanned Systems, Beijing Institute of Technology, Beijing 100081, China (e-mail: jiaqi\_zhu@bit.edu.cn; dengfang@bit.edu.cn).}
\thanks{Shaofeng Cai and Beng Chin Ooi are with the School of Computing, National University of
Singapore, Singapore 117417 (e-mail: shaofeng@comp.nus.edu.sg; ooibc@comp.nus.edu.sg).}
\thanks{Jie Chen is with Harbin Institute of Technology, Harbin 150001, China, and also with the Beijing Institute of Technology, Beijing 100081, China (e-mail: chenjie@bit.edu.cn).}
\thanks{Wenqiao Zhang is with Digital Media Computing \& Design Lab, Zhejiang University, Hangzhou 310000, China (e-mail: wenqiaozhang@zju.edu.cn).}
\thanks{Corresponding author: Fang Deng (dengfang@bit.edu.cn)}
}


\markboth{This paper has been accepted for publication in IEEE Transactions on Pattern Analysis and Machine Intelligence (TPAMI).}
{Jiaqi Zhu \MakeLowercase{\textit{et al.}}: Catching Every Ripple: Enhanced Anomaly Awareness via Dynamic Concept Adaptation}












    \maketitle

\begin{abstract}
\pamirevise{
Online anomaly detection (OAD) plays a pivotal role in real-time analytics and decision-making for evolving data streams.
However, existing methods often rely on costly retraining and rigid decision boundaries, limiting their ability to adapt both effectively and efficiently to concept drift in dynamic environments.
To address these challenges, we propose \name, a dynamic concept adaptation framework for OAD that unifies on-the-fly parameter shifting and dynamic thresholding within a single online paradigm.
\name first learns a static detector on historical data to capture recurring \textit{central concepts}, and then transitions to a dynamic mode to adapt to \textit{new concepts} as drift occurs.
Specifically, \name employs a novel dynamic concept adaptation mechanism that leverages a hypernetwork to generate instance-aware parameter shifts for the static detector, thereby enabling efficient and effective adaptation without retraining or fine-tuning.
To achieve robust and interpretable adaptation, \name introduces a lightweight evolution controller to estimate instance-level concept uncertainty for adaptive updates.
Further, \name employs a dynamic threshold optimization module to adaptively recalibrates the decision boundary by maintaining a candidate window of uncertain samples, which ensures continuous alignment with evolving concepts.}
Extensive experiments demonstrate that \name significantly outperforms existing OAD approaches across a wide spectrum of application scenarios.

\end{abstract}




\section{Introduction}\label{sec:introduction}

\IEEEPARstart{A}{nomaly} detection (AD)
\pamirevise{plays a pivotal role across various domains by uncovering hidden irregularities that deepen data understanding, enable timely interventions, and support reliable decision-making in complex, data-driven systems~\cite{qiu2024self,lu2024robust,zhu2025context,zhu2023meter,xiang2024exploiting}.
While extensive progress has been made in detecting anomalies in static data settings~\cite{kim2020rapp,lai2019robust,yoon2022adaptive,mirsky2018kitsune,zhu2022gaussian}, extending AD to streaming environments remains substantially more challenging.
This task, known as \textit{online anomaly detection (OAD)}, aims to identify anomalous instances in real time from evolving data streams to preserve data integrity and ensure operational reliability across critical applications.
}

\pamirevise{In OAD tasks, \textit{concept drift} poses the central challenge, as unpredictable shifts in the underlying data distribution or correlation structure undermine the validity of models trained on historical data over time.
While traditional static AD approaches may perform well within specific periods, they quickly lose efficacy once concept drift occurs, failing to capture newly emerging abnormal behaviors.
In this context, an anomaly is defined relative to the current nominal distribution of the data stream, which evolves over time.
A pertinent example arises in the financial domain, where transaction patterns evolve due to market trends, seasonal fluctuations, or new trading strategies.
This requires that fraudulent activities be identified relative to the nominal distribution.
Since anomalies are inherently context-dependent, a score that indicates normality under one concept may correspond to an anomaly under another,
rendering static thresholds ineffective.}

\pamirevise{
Unlike concept drift detection, which focuses on identifying when distributional changes occur, OAD under concept drift operates at a finer granularity by continuously discerning rare instance-level deviations within the evolving nominal distribution.
Notably, a concept change point itself is not necessarily anomalous. For instance, a shift from daily to holiday transactions in financial monitoring redefines the nominal concept, whereas anomalies are deviations within this new concept, such as fraudulent transactions concealed among legitimate holiday purchases.
This distinction highlights OAD's unique challenges: beyond merely signaling drift, it requires maintaining sensitivity to subtle, context-dependent anomalies amid continuously evolving concepts.
}

\begin{figure}[t]
    \centering
    \includegraphics[width=1\linewidth]{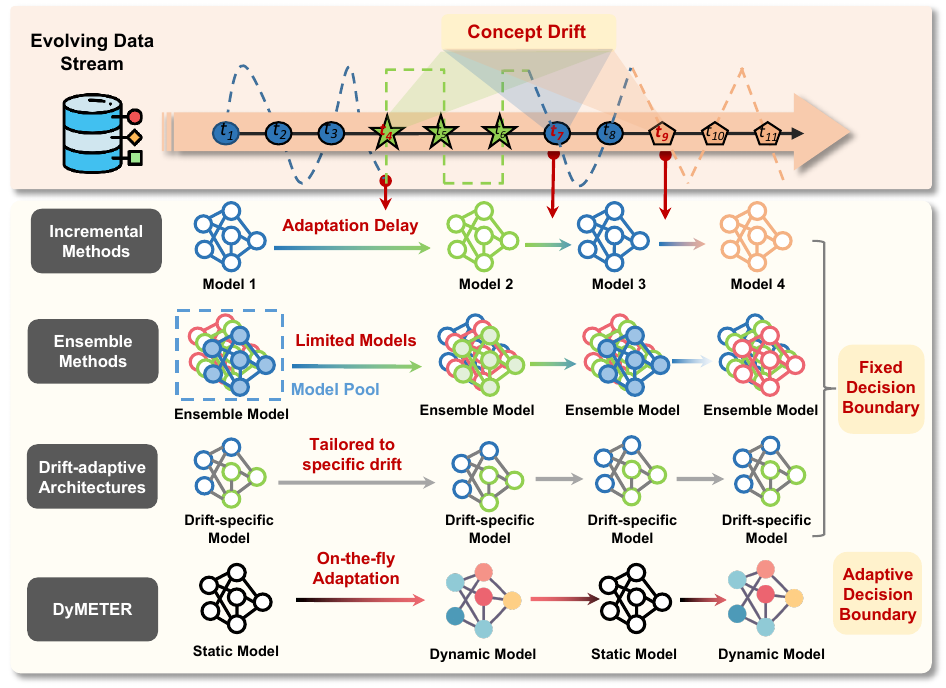} \vspace{-4mm}
\caption{\pamirevise{Adaptation paradigms of \name and existing approaches for handling concept drift in evolving data streams.}}
    \label{fig:intro}
    \vspace{-4mm}
\end{figure}

\pamirevises{
Recently, several preliminary attempts have been proposed to support OAD in evolving data streams, as depicted in Figure~\ref{fig:intro}.
\emph{Incremental methods}~\cite{bhatia2022memstream, bhatia2021mstream,yoon2021multiple, guha2016robust,boniol2021sand} typically construct an initial model that is gradually updated as new data arrives.
However, they often rely on repeated retraining or fine-tuning, resulting in high computational cost and slow responsiveness to emerging concepts.
\emph{Ensemble methods}~\cite{yoon2022adaptive,gopalan2019pidforest,kieu2019outlier} maintain multiple models tailored to different concepts, but their effectiveness is constrained by the size and diversity of the model pool, while sustaining and updating multiple models incurs substantial computational and memory overhead.
More recently, \emph{drift-adaptive architectures}~\cite{wang2023drift,dai2024sarad,li2024state,kim2024model} embed adaptation mechanisms directly into model design. Nevertheless, they are typically tailored to specific drift patterns and thus lack flexibility across diverse drift scenarios.
%
Despite effectiveness in specific settings, existing approaches share several fundamental challenges: (i) adaptation is typically driven by coarse window or batch-level signals rather than instance-level control, (ii) model adaptation often requires continuous tuning or maintaining multiple models instead of efficient inference-time adaptation, and (iii) decision boundaries are typically fixed, making them insensitivity to subtle context-dependent anomalies under concept drift.
}

\pamirevises{
To address these challenges, we propose \name, a novel online anomaly detection framework that reframes adaptation under concept drift as instance-aware, inference-time evolution with dynamic decision boundary calibration.
The core insight behind \name is that historical data often captures the dominant concept patterns and thus comprises the recurring \textit{central concepts}, while newly arriving data typically deviate from these patterns in a gradual and localized manner.
Guided by this understanding, instead of incremental updating, ensemble modeling, and drift-specific heuristics, \name anchors detection to these central concepts and adapts to \textit{emerging concepts} on the fly by jointly evolving the detector's behavior and its decision boundary during online detection as data streams evolve.}


\pamirevises{
Specifically, \name operationalizes dynamic concept adaptation by directly addressing the fundamental challenges of existing approaches.
First, to enable principled instance-level control, \name models concept uncertainty derived from the evidential deep learning (EDL) theory~\cite{sensoy2018evidential} for each incoming instance, providing a continuous and interpretable signal to determine when adaptation is necessary.
Second, to achieve efficient concept adaptation, \name performs inference-time model evolution by dynamically generating instance-conditioned parameter shifts via a hypernetwork. This allows the detector to rapidly re-center toward emerging concepts in a fine-grained and responsive manner, without inefficient gradient-based optimization.
Finally, to maintain reliable anomaly decisions under evolving concepts, \name incorporates cost-sensitive adaptive thresholding that jointly accounts for recent score statistics and concept uncertainty, enabling continuous decision boundary calibration.
As a result, \name attains a unified capability for instance-aware model evolution and dynamic decision calibration, effectively mitigating the coarse adaptation control, high adaptation cost, and decision miscalibration that commonly limit existing OAD methods under concept drift.

We summarize our main contributions as follows:
}

\ignore{
The core idea of \name is to recognize when a static detector becomes unreliable under concept drift, dynamically adapt its parameters, and recalibrate its decision boundary to remain aligned with the evolving concept.
To this end, \name introduces an Intelligent Evolution Controller (IEC) that evaluates each incoming instance by estimating concept uncertainty with evidential deep learning (EDL) theory~\cite{ng2011dirichlet,sensoy2018evidential}.
When drift-affected instances are identified, the framework transitions from the Static Concept-aware Detector (SCD) to the Dynamic Shift-aware Detector (DSD), where a lightweight hypernetwork generates \textit{parameter shifts} conditioned on the instance's features during online inference.
This mechanism re-centers the detector toward the new concept in a timely, instance-aware manner, without costly retraining or fine-tuning.
In parallel, \name incorporates Dynamic Threshold Optimization (DTO) to recalibrate the decision boundary online using recent score statistics and regularized by high-uncertainty points, yielding an adaptive threshold that reduces false alarms while preserving sensitivity to true anomalies.
Furthermore, the EDL-based IEC enhances interpretability by linking low Dirichlet concentration to weak evidential support in the detector's decision, enabling principled uncertainty modeling and transparent anomaly detection.
}

\begin{itemize}[leftmargin=*]
    \item \pamirevises{We present \name, a novel OAD framework that unifies inference-time adaptation and dynamic decision calibration to robustly handle concept drift in evolving data streams.}

    \item \pamirevises{We incorporate a lightweight evolution controller into \name that enables evidential concept uncertainty modeling, supporting fine-grained adaptation control and interpretable anomaly detection.}
    
    \item \pamirevises{We propose a dynamic concept adaptation mechanism that efficiently re-centers the model toward evolving concept by generating instance-aware parameter shifts via a hypernetwork, avoiding continuous retraining or finetuning.
    }
    
    \item \pamirevises{Extensive experiments demonstrate that \name consistently outperforms state-of-the-art OAD baselines across diverse drift scenarios.}
    
\end{itemize}

The remainder of this paper is organized as follows: Section~\ref{sec:preliminary} introduces the preliminaries, including key concepts and background information.
Section~\ref{sec:methodology} presents our \name framework, providing a comprehensive overview of its modules, optimization schemes, and discussions on its key characteristics.
In Section~\ref{sec:experiment}, we showcase experimental results on various benchmarks.
\pamirevise{Section~\ref{sec:related work} reviews related work in the field. Section~\ref{sec:limitations} outlines limitations and future directions, and finally, Section~\ref{sec:conclusion} concludes the paper.}

\section{Preliminaries}
\label{sec:preliminary}




\subsection{Problem Formulation}

\pamirevise{\emph{Anomaly detection} aims to identify data instances that diverge from established norms.
Specifically, this paper focuses on unsupervised \emph{online anomaly detection} (OAD), which seeks to identify anomalies within live data streams without labeled information, as delineated in Definition~\ref{def:OAD}.
The task becomes particularly challenging under \emph{concept drift}, a phenomenon where the underlying statistical or distributional attributes of a dataset evolve over time, as described in Definition~\ref{def:concept drift}.
Such dynamics make it difficult to maintain reliable anomaly detection in real-world scenarios.}

\ignore{
\blue{The objective of \textit{Anomaly detection} is to identify data samples that deviate from the majority or exhibit unusual patterns.
In particular, the focus of this paper is on \textit{online anomaly detection} (OAD), which is to detect anomalies in data streams in real-time, as formulated in Definition~\ref{def:OAD} below.
Further, our framework focuses on unsupervised anomaly detection in data streams, where no labeling information is available.}

  
\blue{In real-world scenarios, data is often dynamic and subject to constant changes~\cite{lu2018learning}, a phenomenon known as \emph{concept drift}, where the statistical or distributional properties of the data within a certain domain change over time, as formally defined in Definition~\ref{def:concept drift}. }
This change in distribution can lead to a significant drift in the patterns and relationships of the data, presenting challenges in detecting anomalous events. }

\blue{
\begin{definition}[Online Anomaly Detection]
    \label{def:OAD}
    \pamirevise{
    Given an incoming data stream $\mathcal{X} =\left\{ \vec{x}_1, \dots, \vec{x}_t, \dots \right \} $, where each instance $ \vec{x}_t=\left( x_{t1},\dots,x_{td} \right)$ is a multivariate feature vector, \textit{online anomaly detection} aims to determine whether $\vec{x}_t$ is anomalous or not at each time step $t$.}\vspace{-2mm}
\end{definition}}

\blue{
\begin{definition}[Concept Drift]
    \label{def:concept drift}
    Concept drift occurs at time step $t$ when the underlying joint probability $P\left( \vec{x}, y \right)$ of input data $\vec{x}$ and its corresponding label $y$ changes at time $t$, such that, $P \left( \vec{x}_t,y_t \right) \ne P\left( \vec{x}, y \right)$.
\end{definition}
}


\pamirevise{In practice, the input instance $\vec{x}_t$ comprises $d$ attributes that can be either categorical or numerical depending on the application domain.
For example, in financial transaction monitoring, $\vec{x}_t$ characterizes
a transaction at time $t$ with dimensions such as amount, merchant type, and geolocation.
As these attributes evolve, concept drift can cause the same anomaly score to indicate normal behavior under one concept but an anomaly under another, such as high-amount transactions being typical during holiday seasons yet anomalous in routine periods.
Therefore, it is crucial for the model to grasp the evolving dynamics of the data stream and adapt to concept drifts effectively over time.}
\pamirevises{This involves dynamically generating an anomaly score from a detection model for each data instance $\vec{x}_t$, assessing its deviation from normal behavior within the stream's varying concepts.
A sample is then identified as anomalous if its anomaly score surpasses an adaptive threshold, which continuously refines the detection criteria to remain aligned with evolving data characteristics.}
\pamirevises{Table~\ref{tab:notation} summarizes the key notations used throughout the paper to facilitate quick reference and improve readability.}


\subsection{Hypernetwork}
Hypernetworks function as a meta neural network that dynamically produces the parameters for a secondary \textit{primary network}, enabling it to adjust its parameters based on specific inputs.
\pamirevise{In our framework, a hypernetwork modifies the parameters of our base model, serving as the primary network to respond to evolving concepts in the data stream.}
\pamirevises{Specifically, we evaluate the parameter changes within our base model, which consists of $N_l$ layers.
For each layer, the hypernetwork assigns a parameter matrix $K^{(n)}\in \mathbb{R}^{N_{in}\times N_{out}}$, where $N_{in}$ and $N_{out}$ denote the input and output dimensionalities of the layer, respectively.
This process is conceptualized as follows:}
\begin{align}
    K^{(n)} = \xi(\vec{r}^{(n)};\Theta_h), \forall n=  1, \cdots, N_l
\label{eq:hyper}
\end{align}
where $\vec{r}^{(n)}$ is a random vector, and $\xi(\cdot)$ denotes a randomly initialized multi-layer perception (MLP) parameterized by $\Theta_h$.
This setup allows for backpropagation to $\vec{r}^{(n)}$ and $\xi(\cdot)$, ensuring the primary network adapts effectively to the data stream through end-to-end training.

\subsection{Evidential Deep Learning}
\label{sec:edl}
\textit{Evidential Deep Learning} (EDL) ~\cite{sensoy2018evidential} advances a probabilistic framework within deep learning by treating neural network outputs as distributions over potential class probabilities, specifically by applying a Dirichlet prior to these probabilities.
This methodology extends beyond providing mere point predictions by offering uncertainty estimates for each output.
Concretely, in typical C-class classification tasks, a deep neural network (DNN) uses a softmax function to transform instance $\vec{x}$ outputs into class probability vectors $\vec{p}$.
EDL enhances this by applying a Dirichlet distribution over $\vec{p}$, effectively modeling each class's probability distribution for $\vec{x}$.
The Dirichlet distribution, formulated as:
\begin{equation}
\begin{aligned}
\scalebox{0.95}{$P(\vec{p}|\vec{x};\Theta_e)=Dir(\vec{p}|\vec{\alpha})=
 \begin{cases} 
\frac{\Gamma (\sum_{c=1}^C \alpha_{c})}{\prod_{c=1}^C\Gamma(\alpha_{c})}\prod_{c=1}^C p_{c}^{\alpha_{c}-1}, \!\! \!\! & \mbox{if } \vec{p} \in \Delta^C\\
\quad\quad\quad\quad0 \quad\quad\quad\quad\quad, &  \!\!\mbox{otherwise}
\end{cases} $}
 \label{eq:edl}
\end{aligned}
\end{equation}
\pamirevises{where $\Theta_e$ denotes the parameters of the evidential classifier that maps an input $\vec{x}$ to Dirichlet evidence.
The Dirichlet distribution is parameterized by $\vec{\alpha}$, which are predicted by the evidential classifier via $\vec{\alpha}=g(f(\vec{x}))$, where $f(\cdot)$ denotes the classifier's output mapping from $\vec{x}$ to evidence, and $g(\cdot)$ is an exponential function enforcing non-negativity.}
$\Gamma(\cdot)$ represents the Gamma function, and $\Delta^C$ is the $C$-dimensional unit simplex, $\Delta^C$=$\{\sum_{c=1}^C p_{c}=1 \ {\rm and} \ 0\leq p_c \leq 1 $\}.
This probabilistic approach allows predictions for $\vec{x}$ to reflect a range of possible outcomes, enriching the model's interpretability and capability to handle uncertainty, particularly valuable for managing concept drift in dynamic environments.
Within the \name framework, EDL plays a vital role in quantifying concept uncertainty, facilitating the model evolution based on the evaluated uncertainty of each input.


\begin{table}[t]
   \small
    \centering
    \renewcommand{\arraystretch}{1.2}
\caption{\pamirevises{Summary of key notations.}}
    \vspace{-2mm}
    \label{tab:notation}
    \resizebox{1\columnwidth}{!}{
        \begin{tabular}{c l}
    \toprule[1.5pt]
 Notation & Description\\
  \toprule[0.8pt]
$\mathcal{X} =\left\{ \vec{x}_1, \dots, \vec{x}_t, \dots \right \}$ & Data stream, where $\vec{x}_t$ denotes the input instance at time $t$. \\
$\Theta_s=(\Theta_s^E,\Theta_s^D)$ & Parameters of the SCD.\\
$\Theta_d=(\Theta_d^E,\Theta_d^D)$ & Parameters of the DSD.\\
$\Delta\Theta_d=(\Delta\Theta_d^E, \Delta\Theta_d^D)$ & Parameter shifts produced by the hypernetwork.\\
$\mathcal{E}_s(\cdot)$, $\mathcal{D}_s(\cdot)$& Encoder and decoder of SCD.\\
$\mathcal{E}_d(\cdot)$, $\mathcal{D}_d(\cdot)$& Encoder and decoder of DSD.\\
$\vec{y}$,$\vec{y'}$ & Reconstructed outputs of SCD and DSD.\\
$\vec{z}$,$\vec{z'}$ & Latent representation vectors of SCD and DSD.\\
$\Theta_c$ & Parameters of the IEC.\\
\pamirevises{$\tilde{l}$} & \pamirevises{Pseudo label for IEC training.}\\
\pamirevises{$\vec{\alpha}$, $\alpha_c$} & \pamirevises{Dirichlet concentration vector and its component for class $c$.}\\
$\mathcal{U}_{ctr}$ & \pamirevises{Concept uncertainty estimated by the IEC.}\\
$\mu_e$ & Uncertainty threshold used to filter unreliable samples.\\
$\mu_p$ & Reconstruction error threshold for pseudo-labeling.\\
$\mathcal{A}(\cdot)$ & \pamirevises{Anomaly score assigned to an input sample.}\\
$\lambda$ & Weight that scales the effect of uncertainty on anomaly score.\\
\pamirevises{$\mathcal{F}_{c}(\cdot)$} & \pamirevises{Expected cost function used for dynamic thresholding.} \\
$\mathcal{W}_N$, $\mathcal{W}_C$ & Sliding windows for normal and uncertain samples.\\
\pamirevises{$r_e$} & \pamirevises{Reference reconstruction error for anomaly score scaling.}\\
$\mu_a^*$ & Final adaptive decision threshold for anomaly detection.\\
$\tau$, $\kappa$ & Hyperparameters for quantile level and regularization. \\
$\mu_o$ & Threshold for triggering offline update.\\
    \bottomrule[1.5pt]
        \end{tabular}
        }\vspace{-4mm}
\end{table}

\section{Methodology}
\label{sec:methodology}

\begin{figure*}[t]
    \centering 
    \includegraphics[width=1\linewidth]{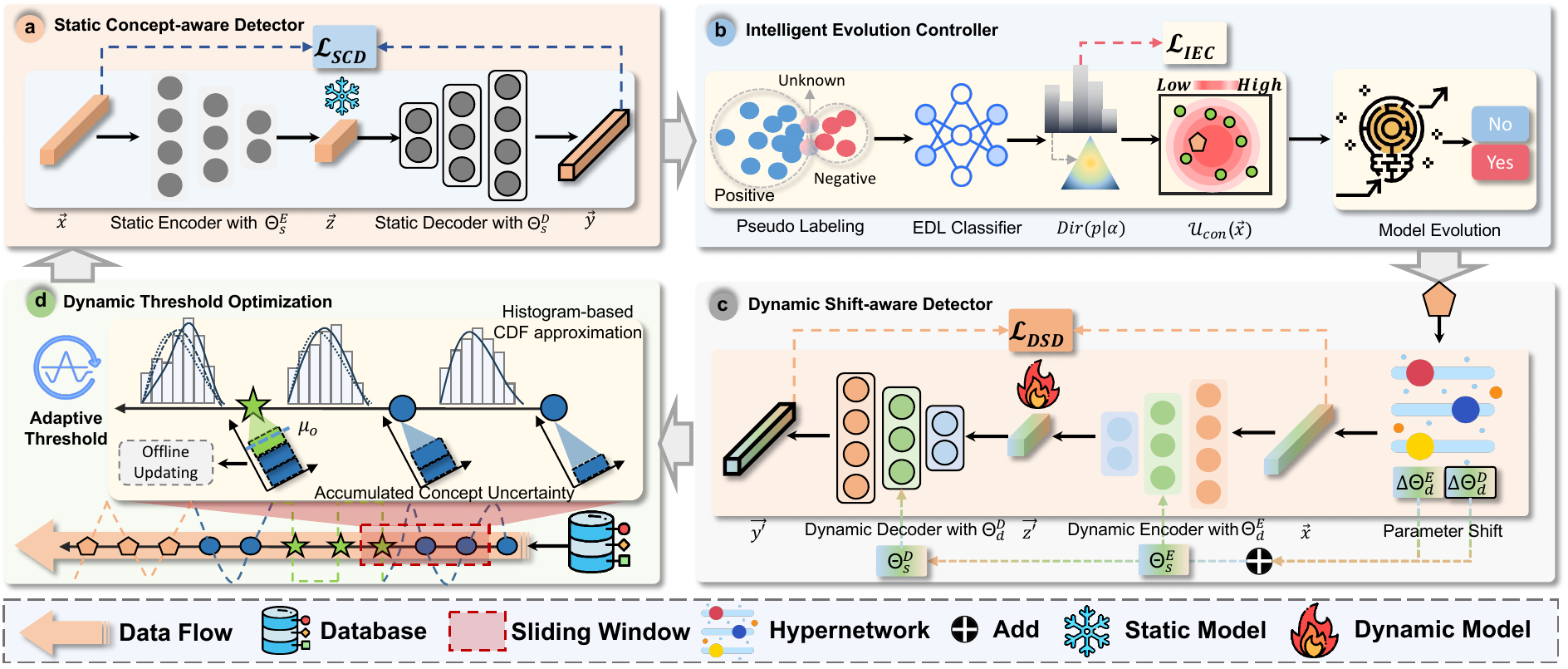}
    \caption{Overview of the proposed \name. (a) Static Concept-aware Detector (SCD) is trained on historical data to model the central concepts.
    (b) Intelligent Evolution Controller (IEC) timely measures the concept uncertainty to determine the necessity of model evolution.
    (c) Dynamic Shift-aware Detector (DSD) dynamically updates SCD with the instance-aware parameter shift by considering the concept drift.
    (d) Dynamic Threshold Optimization (DTO) establishes a cost-sensitive decision boundary by tracking the distribution of reconstruction errors, followed by an accumulated concept uncertainty-based updating strategy.}
    \label{fig:framework}
    \vspace{-1mm}
\end{figure*}

\subsection{\pamirevises{Overview of \name}}
\label{sec:overview}
\pamirevises{
DyMETER is a dynamic concept adaptation framework for online anomaly detection under evolving concepts, enabling instance-aware adaptation of both detector parameters and decision boundaries at inference time.
At a high level, \name comprises four tightly coupled components. The Static Concept-aware Detector (SCD) learns central concepts from historical data and provides anomaly evidence for the Intelligent Evolution Controller (IEC), which estimates concept uncertainty to determine whether an incoming instance aligns with established concepts or signals potential evolution. Guided by this uncertainty, the Dynamic Shift-aware Detector (DSD) selectively adapts detector parameters at the instance level, while the Dynamic Threshold Optimization (DTO) continuously calibrates the anomaly decision boundary to ensure robustness under evolving data streams.
Through this coordinated design, \name forms a unified uncertainty-guided adaptation framework for robust and responsive online anomaly detection under evolving concept drifts.
}


\subsection{Static Concept-aware Detector}
\label{sec:scd}
The Static Concept-aware Detector (SCD) is designed to identify anomalies within the central concepts of a data stream by capturing the overall distribution of historical data.
The SCD uses an autoencoder, known for its effectiveness and unsupervised nature~\cite{ma2024rethinking,abdulaal2021practical}, as the static detector to effectively identify predominant anomaly types within the central concepts of unlabeled data streams.


\noindent\textbf{Static Detector.}
The workflow of our static detector operates in two primary phases: encoding and decoding.
\pamirevises{Concretely, given an input instance $\vec{x}$ in the data stream $\mathcal{X}$, the encoder and decoder jointly form an autoencoder-based detector with parameters $\Theta_s=(\Theta^E_s,\Theta^D_s)$.
The encoder $\mathcal{E}_s(\cdot;\Theta^E_s)$ compresses the representation of the input $\vec{x}$ into a latent vector $\vec{z}$, and subsequently the decoder $\mathcal{D}_s(\cdot;\Theta_s^D)$ reconstructs the original input using the latent vector $\vec{z}$ as follows:}
\begin{align}
\pamirevise{
    \vec{x} \Rightarrow \mathcal{E}_s(\vec{x};\Theta_s^E)=\vec{z} \Rightarrow  \mathcal{D}_s(\vec{z};\Theta_s^D)=\vec{y}
}
    \label{eq:scd}
\end{align}

The static detector is initially trained on a small subset of the historical data stream $\mathcal{X}_h$, which calculates the reconstruction error by measuring the squared deviations between each input and its reconstructed output, denoted by $L_2(\cdot)$, formulated as:
\begin{align}
    \mathcal{L}_{SCD}(\Theta_s)=L_2(\vec{x}, \vec{y})=\frac{\sum_{i=1}^{n} (\vec{x}_i-\vec{y}_i)^2}{n}\,,
\label{eq:loss_scd}
\end{align}
where $n$ denotes the dimension of each input instance.
By constraining the model to internalize essential relationships and features of the input instances via $\mathcal{L}_{SCD}$, the autoencoder-based static detector functions effectively as an identity mapper for its training distribution.
This setup allows it to reliably regenerate unseen samples from $\mathcal{X}_h$, where anomalies are revealed through elevated reconstruction errors.


\subsection{Intelligent Evolution Controller}
\label{sec:isc}
The reconstruction error from the static detector serves as an indicator of normality or anomaly in the input data. However, its reliability decreases with out-of-distribution data, particularly under conditions of concept drift~\cite{graham2023denoising,haroush2021statistical}.
To address this challenge, we propose an Intelligent Evolution Controller (IEC), which is designed to evaluate whether the input instance aligns with the established central concepts or represents emerging new concepts.
Furthermore, the IEC enables dynamic adjustments to the static detector by timely monitoring the uncertainty of the prediction.

\noindent\textbf{Pseudo Labeling Strategy.}
IEC is a lightweight evidential classifier with parameters $\Theta_c$ trained with the high-confidence pseudo labels \pamirevises{$\tilde{l}$} generated by the static detector.
We categorize poorly reconstructed inputs with a label of $1$, and well reconstructed inputs with a label of $0$. 
This labeling strategy allows the IEC to discern whether incoming data matches the established central concepts in the SCD or unseen patterns.
\begin{equation}
\begin{aligned}
\scalebox{0.91}{$
\!\!\!\!\!\! \pamirevises{\tilde{l}}(\vec{x})=
 \begin{cases} 
1 (\texttt{Positive}), \!\! & \mbox{if } \pamirevises{L_{2}(\vec{x},\vec{y})} >\mu_p \mbox{ and } \mathcal{U}_{ctr}(\vec{x})\leq \mu_{e}\\
0 (\texttt{Negative}), \!\! & \mbox{if } \pamirevises{L_{2}(\vec{x},\vec{y})} \leq \mu_p \mbox{ and } \mathcal{U}_{ctr}(\vec{x})\leq \mu_{e}\\
\textbf{-}(\texttt{Unknown}), & { \mathcal{U}_{ctr}(\vec{x} ) > \mu_{e}}
\end{cases} $}
\label{eq:pseudo_label}
\end{aligned}
\end{equation}
\noindent where \pamirevises{$L_{2}(\vec{x},\vec{y})$} is the reconstruction error of the input instance given the static detector computed by taking the root mean squared error between $\vec{x}$ and the reconstructed output $\vec{y}$. 
$\mu_p$ is a predefined pseudo-labeling threshold determined by setting a proportion of the sorted reconstruction error over all training samples.
$\mathcal{U}_{ctr}(\vec{x})$ is the concept uncertainty that determines the necessity of model evolution, which will be introduced in detail next.
$\mu_e$ is a predefined threshold to ensure that the static detector is only trained with high-confidence samples with a low concept uncertainty $\mathcal{U}_{ctr}(\vec{x})$, \emph{i.e.}, $\vec{x}$ is not involved in training if the $\mathcal{U}_{ctr}(\vec{x}) > \mu_{e}$.

\pamirevise{\noindent\textbf{Concept Uncertainty Estimation.}
We introduce the predictive probability of the evidential learning model following Eq.~(\ref{eq:edl}).}
\pamirevises{Given the IEC parameterized by $\Theta_c$, for a given instance $\vec{x}$, the Dirichlet‐based predictive probability for class $c$, denoted as $\hat{p}_c=\hat{P}(y=c|\vec{x};\Theta_c)$, is obtained by marginalizing over $\vec{p}$ as:}
\begin{equation}
\begin{aligned}
\hat{P}(y=c|\vec{x};\Theta_c)&= \int P(y=c|\vec{p};\Theta_c)P(\vec{p}|\vec{x};\Theta_c)d\vec{p}\\
&=\frac{\alpha_c}{\sum_{k=1}^C\alpha_k}=\pamirevises{\mathbb{E}_{\vec{p}\sim \mathrm{Dir}(\vec{\alpha})}[p_c]}\,,
\label{eq:prediction}
\end{aligned}
\end{equation}
\noindent
\pamirevises{where $C$ denotes the number of classes, set to two in anomaly detection setting.
$\vec{p}$ denotes the class probability vector modeled as a Dirichlet random variable, with $p_c$ its component for class $c$, and $\vec{\alpha}$ denotes the corresponding concentration vector with component $\alpha_c$ produced by the evidential classifier as described in Section~\ref{sec:edl}.}

\pamirevises{IEC is trained by pseudo labels from central concepts in Eq.~(\ref{eq:pseudo_label}).
From a Bayesian perspective, its objective can be interpreted as the negative log marginal likelihood obtained by marginalizing class probabilities under a Dirichlet prior.}
To address the class imbalance issue inherent in anomaly detection, we introduce a focal term that places greater emphasis on minority and misclassified examples.
Concretely, our IEC objective is given by:
\begin{align}
    \mathcal{L}_{IEC}(\Theta_c)=(1-\hat{p}_c)^\gamma({\rm log}(\sum_{k=1}^C \alpha_k)-{\rm log} \alpha_c)\,,
\label{eq:loss_iec}
\end{align}
\pamirevises{where $\gamma\geq0$ is the focal exponent and $\hat{p}_c$ is the predictive probability for class $c$ as given by Eq. (\ref{eq:prediction}), which is binary in anomaly detection setting.}
\pamirevise{
However, when the concepts of current samples differ markedly from the established central concepts, the evidence supporting these samples becomes insufficient, as the controller now lacks the information about such new concepts~\cite{sensoy2018evidential,ng2011dirichlet}.
To quantify this deviation, we propose \textit{concept uncertainty} $\mathcal{U}_{ctr}$ defined as:}
\begin{equation}
\begin{aligned} 
&\mathcal{U}_{ctr}(\vec{x};\Theta_c) = \sum_{c=1}^C \hat{P}(y=c|\vec{x};\Theta_c) \left(\Phi(\alpha_{c}+1\right)- \\& \pamirevises{\Phi(\sum_{c=1}^C \alpha_{c}+1))}-\sum_{c=1}^C \hat{P}(y=c|\vec{x};\Theta_c) {\rm log} \pamirevises{\hat{P}(y=c|\vec{x};\Theta_c)}\,,
\label{eq:uncer}
\end{aligned}
\end{equation}
where $\Phi(\cdot)$ is the digamma function and we utilize mutual information as a metric to evaluate the dispersion of the Dirichlet distribution across the simplex following~\cite{sensoy2018evidential}.

\pamirevise{
Theoretically, the \pamirevises{Dirichlet concentration vector $\vec{\alpha}$} quantifies the amount of evidence supporting each class prediction. When an input $\vec{x}$ conforms to the previously learned concept, the corresponding Dirichlet distribution remains concentrated, yielding low mutual information and a small $\mathcal{U}_{ctr}(\vec{x})$. In contrast, when $\vec{x}$ arises from a shifted concept, the evidence becomes diffuse and uncertain, increasing both the entropy of the predictive distribution and the mutual information term.
Consequently, $\mathcal{U}_{ctr}$ rises proportionally to the model’s lack of knowledge about the new concept.
When $\mathcal{U}_{ctr}(\vec{x}) > \mu_e$, the static detector's prediction is deemed unreliable, prompting the dynamic shift-aware detector to reassess the input $\vec{x}$, thereby providing a theoretically grounded mechanism for identifying and responding to concept drift.
}

\subsection{Dynamic Shift-aware Detector}
\label{sec:dsd}
\pamirevise{
Recall that the SCD captures central concepts from historical data but struggles to model evolving data properties accurately once concept drift occurs.
Therefore, it is crucial to promptly capture emerging concepts to sustain detection accuracy over time.
To this end, we introduce the Dynamic Shift-aware Detector (DSD), which enhances the SCD by enabling it to adapt to new concepts in streaming data dynamically.
}

\noindent\textbf{Parameter Shift Measurement.}
\pamirevise{
We design a hypernetwork to adapt the key parameters of the static detector in \name under concept drift, denoted collectively as $\Delta\Theta_d=(\Delta\Theta_d^E, \Delta\Theta_d^D)$, where $\Delta\Theta_d^E$ and $\Delta\Theta_d^D$ represent the encoder and decoder parameter shifts, respectively.
}
Unlike the traditional hypernetwork as shown in Eq.~(\ref{eq:hyper}) that uses a randomly initialized vector $\vec{r}^{(n)}$ to generate the model parameters, ignoring the interaction with the input instance, we replace $\vec{r}^{(n)}$ with representations derived directly from the current input to incorporate instance-specific adjustments.
Concretely, the hypernetwork processes input $\vec{x}$ through a specialized subnetwork $E^{(n)}(\cdot)$ to extract relevant features ${\vec{e}}^{(n)}$ for adjusting parameters of each layer. 
Efficiency is achieved by using a shared encoder network $E_{\rm{share}}(\cdot)$ across all layers, with distinct linear layers producing the necessary layer-specific vectors ${\vec{e}}^{(n)}$.
\begin{align}
\label{eq:lightweight_encoder}
    {\vec{e}}^{(n)} = E^{(n)}(\vec{x}) = {L}^{(n)}_{\rm{layer}}({E}_{\rm{share}} (\vec{x})), \;\rm{n }= 1, \cdots,\mathcal{N}_d,
\end{align}
\noindent
where $N_d$ is the number of layers of the encoder and decoder,
and ${L}^{(n)}_{\rm{layer}}(\cdot)$ is the linear layer that transforms the output of ${E}_{\rm{share}}(\cdot)$ to the features of the $n$-th layer.
The adjustment for the $n$-th layer's parameters in the SCD is represented by matrix  $K^{(n)} \in \mathbb{R}^{N_{in}\times N_{out}}$, correlating to the count of input and output neurons respectively.
This transformation utilizes the feature vector ${\vec{e}}^{(n)}$, which is specific to the input, to modify the parameters of the relevant layer.
In detail, this process involves the use of two MLP layers within the hypernetwork to craft these parameter changes for the $n$-th layer:
\begin{align}
\label{eq:kernal_generation_detail}
{W}^{(n)} &= (W_1{\vec{e}}^{(n)} + \vec{b}_1)W_2 + \vec{b}_2 \\
K^{(n)} &= {W}^{(n)} + {{\vec{b}}}^{(n)}
\end{align}
\noindent
where $W_1$ and $W_2$ are weights of the MLP layers of the hypernetwork, $\vec{b}_1$, $\vec{b}_2$ and  $\vec{b}$ are the biases.

\begin{algorithm}[t]
\small
\pamirevise{
\caption{\name~Training}
\label{alg:pseudo_code_training}
\Input{Small subset of historical stream data $\mathcal{X}_h$}
\Output{\name\ parametrized by $\Theta_s$, $\Theta_c$ and $\Theta_d$}
\Initialization{Randomly initialized $\Theta_s$, $\Theta_c$ and $\Theta_d$}

\Repeat{Convergence}{
  Randomly sample a minibatch\;
  \eIf{SCD not trained in \name}{
    Update $\Theta_s$ via minimizing $\mathcal{L}_{SCD}$ using Eq.~(\ref{eq:loss_scd})\;
  }{
    Pseudo-labeling samples using Eq.~(\ref{eq:pseudo_label})\;
  }
}
\Return{SCD with parameters $\Theta_s$}\;

\Repeat{Convergence}{
  Randomly sample a minibatch\;
  Update $\Theta_c$ via minimizing $\mathcal{L}_{IEC}$ using Eq.~(\ref{eq:loss_iec})\;
  
  Update parameters $\Theta_s$, $\Theta_c$, and $\Theta_d$ via minimizing
  $\mathcal{L}_{DSD}$ using Eq.~(\ref{eq:loss_dsd})\;
}
\Return{\name\ with parameters $\Theta_s$, $\Theta_c$ and $\Theta_d$}\;
}
\end{algorithm}

\ignore{
\begin{algorithm}[t]
\small
\renewcommand{\algorithmicrequire}{\textbf{Input:}}  
\renewcommand{\algorithmicrequire}{\textbf{Initialization:}}

\begin{flushleft}
  \caption{\name Training}
    \textbf{Input}: 
    Small subset of historical stream data $\mathcal{X}_h$ \\
    \textbf{Output}: \name parametrized by $\Theta_s$, $\Theta_c$ and $\Theta_d$\\
    \textbf{Initialization}: Randomly initialized $\Theta_s$, $\Theta_c$  and $\Theta_d$\\
    
     \Repeat {Convergence}{Randomly sample a minibatch\\
        \If{SCD not trained in \name}{Update $\Theta_s$ via minimizing $\mathcal{L}_{SCD}$ using Eq.(\ref{eq:loss_scd})} 
        \Else {
        Pseudo-labeling samples using Eq.(\ref{eq:pseudo_label})
        }
        }        
  \Return{SCD with parameters $\Theta_s$}

   \Repeat {Convergence}{Randomly sample a minibatch\\
    Update $\Theta_c$ via minimizing $\mathcal{L}_{IEC}$ using Eq.(\ref{eq:loss_iec})
    
    Update parameters $\Theta_s$, $\Theta_c$, and $\Theta_d$ via minimizing $\mathcal{L}_{DSD}$ using Eq.(\ref{eq:loss_dsd})
 }
  \Return{\name with parameters $\Theta_s$, $\Theta_c$ and $\Theta_d$}
\end{flushleft} 
\end{algorithm}
}

\noindent\textbf{Dynamic Detector.}
Once the parameter shift of the static detector has been determined, we set the parameters for the dynamic detector $\Theta_d =(\Theta_d^E,\Theta_d^D)$, accordingly:
\begin{equation}
\begin{aligned}
\Theta_d=
 \begin{cases} 
\Theta_d^E =\Theta_s^E+\Delta\Theta_d^E \\
\Theta_d^D=\Theta_s^D+\Delta\Theta_d^D
\end{cases} 
\label{eq:stod}
\end{aligned}
\end{equation}

With the updated parameters for the dynamic detector, the reconstruction process for a given $\vec{x}$ is formulated as:
 \begin{align}
    \vec{x} \Rightarrow \mathcal{E}_d(\vec{x};\Theta_d^E)=\vec{z'} \Rightarrow  \mathcal{D}_d(\vec{z'};\Theta_d^D)=\vec{y'}, \; s.t.\;\;\vec{x} \approx \vec{y'}
\end{align}
where $\vec{z'}$ is the latent representation vector and $\vec{y'}$ is the reconstructed output vector of the dynamic detector.
The training process of the dynamic detector in DSD is similar to that of SCD, with the loss function $\mathcal{L}_{DSD}$ defined as follows:
 \begin{align}
    \mathcal{L}_{DSD}(\Theta_d)=L_2(\vec{x}, \vec{y'})=\frac{\sum_{i=1}^{n} (\vec{x}_i-\vec{y'}_i)^2}{n}
\label{eq:loss_dsd}
\end{align}

\pamirevise{
The training procedure of \name\ is outlined in Algorithm~\ref{alg:pseudo_code_training}.
The SCD is first trained by minimizing $\mathcal{L}_{SCD}$ (Lines 3–4), then the IEC (Line 11) is trained using pseudo labels from the SCD.
The DSD hypernetwork is subsequently optimized with gradients jointly propagated through the SCD and IEC (Line 12) to enable end-to-end adaptation.
}

\subsection{\violet{Dynamic Threshold Optimization}}
\label{sec:dto}
While \name adaptively models the evolving data stream, providing an evolving estimate of reconstruction errors, a fixed threshold for anomaly detection may become misaligned with the data under concept drift.
To address this concern, we introduce a dynamic threshold optimization mechanism that continuously recalibrates the decision boundary for OAD based on recent observations.

\noindent\textbf{\violet{Uncertainty-aware Anomaly Scoring.}}
Let $\mathcal{R}_e$ be the reconstruction error for an incoming data sample $\vec{x}$, produced by the SCD as $L_{2}(\vec{x},\vec{y})$ for static cases or DSD as $L_2(\vec{x}, \vec{y'})$ for concept drift cases.
We establish the anomaly score of \name considering both reconstruction error $\mathcal{R}_e$ and the concept uncertainty $\mathcal{U}_{ctr}$ as:
\begin{align}
\mathcal{A}(\vec{x})= \mathcal{R}_e \exp(\lambda \,\mathcal{U}_{ctr}(\pamirevises{r_{e}}-\mathcal{R}_e))\,,
\label{eq:anomaly_score}
\end{align}
where $\lambda\geq 0$ is the weight controlling how strongly uncertainty calibrates the anomaly score.
\pamirevises{$r_{e}$} is a pivot value representing the referenced reconstruction error.
Specifically, we initialize \pamirevises{$r_{e}$} with the maximum reconstruction error from the training set and update it via an exponential moving average (EMA)~\cite{klinker2011exponential}.
The proposed anomaly scoring mechanism affords a tunable balance, preventing the over-penalization of uncertain, large-error samples that might be normal data in unfamiliar regions and reducing the risk of overlooking uncertain, small-error instances that could be hidden anomalies.

\noindent\textbf{\violet{Cost-sensitive Thresholding.}}
Based on the anomaly scoring, we classify a sample $\vec{x}$ as anomalous if $\mathcal{A}(\vec{x})>\mu_{a}$ and normal otherwise.
However, maintaining the threshold $\mu_{a}$ as a constant proves insufficient in streaming scenarios where the ratio of anomalies may fluctuate when concept drift occurs unpredictably.
In a more comprehensive formulation, one could incorporate the distributions of anomaly score for both normal and anomalous, denoted $p_N(\mathcal{A})$ and $p_A(\mathcal{A})$, with prior probabilities $\pi_N$ and $\pi_A$.
By letting $\mathcal{C}_{fp}$ be the cost of a false positive and $\mathcal{C}_{fn}$ the cost of a false negative, the expected cost function
takes the form as:
\begin{align}
\scalebox{1}{$\mathcal{F}_{c}(\mu_a)=\pi_N \mathcal{C}_{fp} \int_{\mu_a}^{\infty}{p_N(\mathcal{A})}d\mathcal{A}+\pi_A \mathcal{C}_{fn}\int_{0}^{\mu_a}{p_A(\mathcal{A})}d\mathcal{A}\,$},
\end{align}
where the first term measures the contribution from normal data misclassified as anomalies, \emph{i.e.,} false positive rate (FPR), and the second term quantifies the contribution from the undetected anomalies, \emph{i.e.,} false negative rate (FNR).

In principle, minimizing $\mathcal{F}_{c}$ yields an optimal threshold $\mu^{*}_a=\argmin\mathcal{F}_{c}(\mu_a)$ under the specified misclassification costs and prior probabilities.
However, since anomaly samples are often rare in practice, owing to the difficulty in estimating $p_A(\mathcal{A})$, we simplify threshold optimization by focusing on the distribution of reconstruction errors for data that are classified as normal.
This approach is grounded in the fact that normal instances generally dominate the data stream, \emph{i.e.,} $\pi_N \gg \pi_A$,
enabling robust estimation of $p_N(\mathcal{A})$.
Specifically, we maintain a sliding window $\mathcal{W}_N$ consisting of recently classified normal samples.
For each arriving sample classified as normal, we include its anomaly score $\mathcal{A}$ into $\mathcal{W}_N$, and remove the oldest entry when the window reaches a predefined capacity.
An empirical distribution $\hat{p}_N(\mathcal{A})$ of anomaly score is then derived from $\mathcal{W}_N$ and updated incrementally via an empirical cumulative distribution function (CDF).
In particular, a histogram-based approximation of the CDF can be formulated as:
\begin{align}
F_N(\mathcal{A})=\frac{1}{|\mathcal{W}_N|}\sum_{i\in \mathcal{W}_N}{{\mathds{1}}_{(\mathcal{A}_i \leq \mathcal{A})}}
\end{align}
where $\mathds{1}_{(\cdot)}$ denotes the indicator function.
By selecting a preliminary threshold $\mu_a^0$ as the $(1-\tau)$-quantile of $\hat{p}_N(\mathcal{A})$, the system guarantees an upper bound on the FPR as $\tau$.
Nevertheless, controlling the FPR remains challenging when the model displays high uncertainty.
To address this issue, an additional regularization term $\mu_a^r$ is introduced to fine-tune the threshold and mitigate potential misses.
A second sliding window $\mathcal{W}_C$ is thus maintained, containing samples with uncertain anomaly scores close to $\mu_a^0$.
Specifically, the anomaly score of an arriving sample is placed into $\mathcal{W}_C$ if it satisfies both $\mathcal{U}_{ctr}>\mu_{t}$ and $\mathcal{A}\in [\mu_a^0-\delta, \mu_a^0+\delta]$, where the threshold $\mu_{t}$ is set to the maximum concept uncertainty observed across the training set after training and $\delta$ is chosen as the median absolute deviation of $\mathcal{W}_N$.
Let $\widehat{\mathcal{A}}_m$ represent the median anomaly score of samples in $\mathcal{W}_C$.
The regularization term $\mu_a^r$ is then computed by:
\begin{align}
    \mu_a^r=\kappa (\mu_a^0-\widehat{\mathcal{A}}_m)
\end{align}
where $\kappa$ governs the strength of the regularization.
Intuitively, if $\widehat{\mathcal{A}}_m$ is notably less than $\mu_a^0$, then the overall threshold is increased to guard against potential false negatives, conversely, if $\widehat{\mathcal{A}}_m$ is comparable to $\mu_a^0$, then $\mu_a^r$ remains small.
Finally, the threshold for OAD is given by:
\begin{align}
\pamirevise{
\mu_a^*=\mu_a^0+\mu_a^r
}
\label{eq:threshold}
\end{align}

When there is no significant drift, the threshold $\mu_a^*$ is updated incrementally, relying on the estimates of $\hat{p}_N(\mathcal{A})$ and exponentially decaying rank statistics $\widehat{\mathcal{A}}_m$ and $\delta$.
If the IEC signals concept drift, the DSD enhances the static detector to accommodate the new concepts in the current data stream.
At the same time, the threshold optimization procedure is re-initialized to avoid using stale error statistics. Specifically, we (i) discard old entries in $\mathcal{W}_N$ and $\mathcal{W}_C$, (ii) re-estimate $\hat{p}_N(\mathcal{A})$ as new data arrives, and (iii) recompute $\mu^{0}_a=\mathcal{F}_N(\mathcal{A})^{-1}(1-\tau)$ and $\mu_a^r=\kappa (\mu_a^0-\widehat{\mathcal{A}}_m)$ to gain the final threshold $\mu_a^*$.
This ensures that both the underlying model and the decision boundary adapt in tandem, enabling the system to align well with real-world operational constraints while preserving the capacity to adapt to drifting conditions.

\begin{algorithm}[t]
\small
\pamirevise{
\caption{\name Inference}
\label{alg:pseudo_code_inference}
\Input{Evolving stream data $\mathcal{X}$}
{\spaceskip=0.2em plus 0.1em minus 0.1em
\Output{Anomaly scores and decisions for records in $\mathcal{X}$}
}
\Initialization{Trained SCD, IEC and DSD parametrized by $\Theta_s$, $\Theta_c$ and $\Theta_d$}

\For{$\vec{x}$ in $\mathcal{X}$}{
  Compute the concept uncertainty $\mathcal{U}_{ctr}(\vec{x})$ from IEC using Eq.~(\ref{eq:uncer})\;
  \eIf{$\mathcal{U}_{ctr}(\vec{x})>\mu_e$}{
    Escalate to DSD using Eq.~(\ref{eq:stod})\;
    \Return{Anomaly score $\mathcal{A}(\vec{x})$ using Eq.~(\ref{eq:anomaly_score})}\;
  }{
  Inference with SCD using Eq.~(\ref{eq:scd})\;
    \Return{Anomaly score $\mathcal{A}(\vec{x})$ using Eq.~(\ref{eq:anomaly_score})}\;
  }
  Calibrate threshold $\mu_a^*$ using Eq.~(\ref{eq:threshold})\;
 \eIf{$\mathcal{A}(\vec{x})>\mu_a^*$}{
    \Return{\pamirevises{$\vec{x}$ is anomaly}}\;
  }{
  \Return{\pamirevises{$\vec{x}$ is normal}}\;
  }
  Compute $\mathcal{UP}$ using Eq.~(\ref{eq:offline_update})\;
\If{$\mathcal{UP}=1$}{
  Update \name\;
  \textbf{break}\;
}
  }
\Return{\pamirevises{Anomaly score $\mathcal{A}(\vec{x})$ and the decision for each $\vec{x} \in \mathcal{X}$}}\;
}
\end{algorithm}

\noindent\textbf{Offline Updating.}
\label{sec:ous}
The entire framework automatically updates its modules based on the accumulated concept uncertainty within the sliding window $\mathcal{W}_N$ over the evolving data stream.
Specifically, \name keeps monitoring if the concept uncertainty accumulated within this window surpasses a predefined threshold, and frequent occurrence of such events indicates that the SCD trained on the historical data streams is incapable of handling the current data stream due to increased concept drift over time.
This suggests that the entire framework should be updated to align with the current data as follows:
\begin{equation}
\begin{aligned}
\scalebox{0.86}{$\mathcal{UP}=
 \begin{cases} 
1, & \mbox{if } \sum_{i=t}^{t+|\mathcal{W}_N|}({\mathds{1}}_{(\mathcal{U}_{ctr}(\vec{x}_i)>\mu_{e})}  \cdot \mathcal{U}_{ctr}(\vec{x}_i))>\mu_{o}\,\mathrm{or} \ \Delta t>T_{max} \\
0, & \mbox{otherwise }
\end{cases} $}
\label{eq:offline_update}
\end{aligned}
\end{equation}
\noindent
where $\mathcal{UP}$=1 indicates the framework should be updated,
\blue{$\Delta t$} is the time since the last framework update, and $\mu_o$ and $T_{max}$ are the threshold of the framework update and maximum interval since the last update, respectively.
$\sum_{i=t}^{t+|\mathcal{W}_N|}({\mathds{1}}_{(\mathcal{U}_{ctr}(\vec{x}_i)>\mu_{e})}  \cdot \mathcal{U}_{ctr}(\vec{x}_i)$ aggregates the concept uncertainty greater than the threshold $\mu_{e}$ within the current sliding window.
We implement a parallel training approach that simultaneously fine-tunes key modules offline within a sliding window and uses the latest models for OAD, enhancing efficiency and modularity without disrupting the online services.
\pamirevise{
The inference procedure of \name is outlined in Algorithm~\ref{alg:pseudo_code_inference}.
}

\vspace{-2mm}
\subsection{Analysis and Discussion}
\label{sec:analysis_and_discussion}

\pamirevise{
\noindent
\highlight{Effectiveness.}
\name tackles the issue of concept drift by employing a dual-detector framework guided by the IEC.
The static detector utilizes historical data to capture the central concept, which provides a stable foundation for initial anomaly detection, while the dynamic detector adjusts to new concepts by dynamically updating its parameters, ensuring responsiveness to evolving scenarios.
The IEC quantifies uncertainty in the Dirichlet space, where concept drift is observed as increased distributional dispersion when the model encounters an unseen concept.
Unlike entropy-based measures derived from softmax probabilities, which conflate aleatoric and epistemic uncertainty, the evidential formulation explicitly represents how confident the model is in its predictions, thereby enabling timely model adaptation.
Additionally, dynamic threshold optimization within the framework allows for real-time calibration of the decision boundary, which refines the sensitivity of the anomaly detection process.
Building on this, the framework also benefits from an offline updating mechanism.
This further ensures ongoing detection accuracy in real-world applications.
}

\ignore{\name addresses the concept drift challenge by integrating two detectors and an IEC.
The SCD detector leverages historical data and prior knowledge for the detection, while the DSD detector dynamically learns the parameter shift to enhance SCD and adapts to new concepts effectively in an instance-aware manner.
Notably, IEC determines whether concept drift occurs, circumventing the risk of employing an ineffective detection model for anomaly detection.
With IEC, the adaptability and generalizability of \name are substantially improved, leading to enhanced accuracy in anomaly detection.
In addition, the offline update strategy provides an efficient way to keep up with new concepts in evolving data streams, thereby ensuring high-quality detection results.} 

\vspace{-3mm}
\noindent\textbf{\pamirevise{Efficiency.}}
\pamirevise{
Efficiency is critical in OAD for a timely response to evolving data streams.
Leveraging concept uncertainty estimates from EDL, \name signals drift and employs a hypernetwork to dynamically adjust model parameters, thus avoiding frequent retraining or fine-tuning of the full model, and significantly cutting computational costs while enhancing responsiveness.
}
\pamirevise{
The time complexity of \name remains lightweight and well-suited for streaming scenarios. The SCD, implemented as a symmetric autoencoder with $n_S$ layers, incurs a cost of $\mathcal{O}(dd_hn_S)$
for an input $\vec{x}_t \in \mathbb{R}^d$ and latent dimension $d_h<d$.
The IEC is implemented as an $n_I$-layer MLP classifier with complexity $\mathcal{O}(dn_I)$.
For adaptation, the DSD employs an $n_D$-layer hypernetwork that generates per-instance parameter offsets, resulting in $\mathcal{O}(dn_D + |\Theta_s|)$, where $|\Theta_s|$ is the parameter size of the static detector.
Ultimately, the DTO calibrates thresholds using quantile statistics over a sliding window of size $\mathcal{W}_N$, with complexity $\mathcal{O}(\mathcal{W}_N \log \mathcal{W}_N)$, while its uncertainty-based regularization introduces negligible overhead. Overall, \name exhibits linear complexity in the input dimension $d$, with adaptation adding only the cost of generating per-instance parameter offsets of size $|\Theta_s|$, hence ensuring computational efficiency for real-time streaming data.
}

\vspace{-3mm}
\pamirevise{
\noindent
\highlight{Interpretability.}
\name facilitates interpretable anomaly detection by providing trustworthy uncertainty estimates essential for high-stakes scenarios, contrasting with the often unreliable softmax probabilities used by other OAD methods~\cite{yoon2022adaptive,bhatia2022memstream,ntroumpogiannis2023meta,bountrogiannis2022distribution}.
Building on subjective logic (SL) theory~\cite{jsang2018subjective}, \name employs EDL to explicitly model prediction reliability, which forms structured opinions under Dempster–Shafer evidence theory, mapping belief assignments to a Dirichlet distribution for nuanced anomaly detection.
Beyond reliable uncertainty quantification, \name also clarifies the distinction between state changes (i.e., concept drift points) and anomalies.
For example, in financial monitoring, a transition from “daily transactions” to “holiday transactions” constitutes a normal state change, whereas fraudulent transactions hidden among holiday purchases remain anomalies.
\name makes this separation explicit: concept drift is characterized via concept uncertainty estimation by the IEC at the input level to trigger adaptation, while anomalies are identified through reconstruction error-based anomaly scores under the current concept.
This interpretability not only strengthens trust in \name's decisions but also ensures that normal state changes can be effectively tracked without being misclassified as anomalies, while genuine anomalies remain detectable under evolving distributions.
}

\section{Experiments}
\label{sec:experiment}



\subsection{Datasets}\label{sec:dataset}
We utilize 23 benchmark datasets, comprising 19 real-world datasets from various domains and 4 synthetic datasets, designed to simulate different types of concept drift. These datasets, widely referenced in related studies, serve to evaluate the effectiveness of OAD approaches under concept drifts.
We classify these datasets into two OAD settings: continuous, with temporal dependencies between observations, and discrete, with no or unknown dependencies.
Data statistics for each dataset are provided in Table~\ref{tab:dataset_stats}. 
\noindent
\textbf{Real-world datasets.}
We adopt four widely recognized anomaly detection datasets from the UCI repository
and ODDS library~\cite{Rayana:2016}, specifically the Ionosphere (Ion.), Pima, Satellite, and Mammography (Mamm.) datasets.
We also incorporated the BGL dataset~\cite{4273008}, a substantial collection of log messages from a BlueGene/L supercomputer at Lawrence Livermore National Labs, which we processed into a structured data format for analysis.
Additionally, we use popular intrusion detection datasets such as KDDCUP99~\cite{kdd99-web} and NSL-KDD~\cite{tavallaee2009detailed}. 
For time-series analysis, we draw from HexagonML's UCR archive~\cite{UCRArchive2021,wu2021current}, including datasets like Internal Bleeding (I.B., index 133), GaitPhase (index 129), NASA (index 158), EPG (index 145), Activity Recognition (A.R., index 162), and ECG (index 194), known for their robustness and diversity. 
We further include datasets like Machine Temperature and CPU utilization from the Numenta anomaly detection benchmark~\cite{ahmad2017unsupervised}.
Lastly, we integrate the INSECTS dataset~\cite{souza2020challenges}, which consists of optical sensor data collected during the monitoring of flying insects, to simulate concept drift with temperature level as a controlled concept. These varied datasets enabled a thorough examination of detection models under different scenarios and drift conditions.

\pamirevise{\noindent
\textbf{Synthetic datasets.}
We adopt synthetic datasets derived from MNIST~\cite{lecun2002gradient} and FMNIST~\cite{xiao2017fashion}, following the public benchmarks of~\cite{yoon2022adaptive}.
It randomly sets categories as anomaly targets with the remaining serving as normal to simulate concepts.
The occurrence locations of concept drifts are set by replacing the current concept with a new one after a random duration between one and four times the batch size.
Two drift types are considered: abrupt and recurrent, where concepts switch sharply and seen concepts may reappear, and gradual and recurrent, where drifts occur gradually across several batches.
}

\ignore{
Four synthetic datasets~\cite{lecun1998gradient,xiao2017fashion}
are introduced to simulate complex anomaly detection scenarios and data streams~\cite{yoon2022adaptive}. It randomly sets categories as anomaly targets to simulate concepts and sets the duration of each concept randomly to simulate two types of concept drift: "abrupt and recurrent" and "gradual and recurrent".}

\begin{table}[t!]
\small
\centering \vspace{-1mm}
  \renewcommand{\arraystretch}{1}
    \caption{\blue{Dataset statistics.}
    }
    \vspace{-2mm}
    \label{tab:dataset_stats}
    \resizebox{1\columnwidth}{!}{
    \begin{tabular}{c|c|cccrc}
    \toprule[1.8pt]
    \multicolumn{1}{c|}{Scenarios}  &\multicolumn{1}{c|}{Settings}&  Datasets     &   $\#$Obj.   &   $\#$Dim.  &   $\#$Outliers ($\%$)     & Concept drift type  \\ 
    \Xhline{1pt}
    \multicolumn{1}{c|}{\multirow{19}{*}{Real-world}} & \multicolumn{1}{c|}{\multirow{6}{*}{\makecell[c]{Discrete \\ setting}}} & Ionosphere   &   351    &   33      &   126 (35.90$\%$) & Unknown \\
    & & Pima   &   768    &   8    &  268 (34.90$\%$)  & Unknown \\
   & & Satellite   &   6435   &   36   &   2036 (31.64$\%$)& Unknown \\
   & & Mammography   &   11,183     &   6    &   250 (2.32$\%$)    & Unknown\\ 
   & & \revise{BGL}   &  \revise{4,713,493}   &  \revise{ 9}   &  \revise{ 348460 (7.39$\%$)}& Unknown \\
   & & NSL   &  125,973   &   42   &   58630 (46.54$\%$)& Unknown \\
   & & KDD99   &   494,021   &   41   &   97278 (19.69$\%$) & Unknown  \\ \Xcline{2-7}{0.4pt}  
  &  \multirow{12}{*}{\makecell[c]{Continuous \\ setting}} & \violet{Activity Recognition}  &   6,675    &   10   &   59 (0.88$\%$)      & Unknown\\
  &  & \violet{Internal Bleeding} & \blue{7,492} & \blue{10} & \blue{27 (0.36$\%$)} & \blue{Unknown} \\
  &  & \violet{NASA} & \blue{11,299} & \blue{10} & \blue{97 (0.86$\%$)} & \blue{Unknown} \\
  &  & \violet{GaitPhase} & \blue{11,991} & \blue{10} & \blue{45 (0.38$\%$)} & \blue{Unknown} \\
  &  & \violet{EPG} & \blue{30,000} & \blue{10} & \blue{50 (0.17$\%$)} & \blue{Unknown} \\
  &  & \violet{ECG} & \blue{80,000} & \blue{10}& \blue{250 (0.31$\%$)}  & \blue{Unknown} \\
  &  & Machine temperature &   22,695    &   10   &   2268 (10.00$\%$)      & Unknown\\
  &  & CPU utilization  &   18,050    &   10   &   1499 (8.30$\%$)  & Unknown  \\
 \Xcline{3-7}{0.4pt} 
 &   & INSECTS-Abr   &   44,569     &   33      &  529 (1.19$\%$) & Abrupt   \\ 
  &  & INSECTS-Inc   &   48,086    &   33      &   571 (1.19$\%$)  & Incremental    \\ 
 &   & INSECTS-IncGrd   &   20,367    &   33      &    242 (1.19$\%$) & Incremental/gradual  \\ 
 &   & INSECTS-IncRec   &   67,455     &   33    &   800 (1.19$\%$)   & Incremental/recurrent \\ 
      \midrule[1pt]
   \multirow{4}{*}{Synthetic}& \multirow{4}{*}{\makecell[c]{Discrete \\ setting}} & SynM-AbrRec     &   20,480     &   784      &    196 (0.96$\%$)   & Abrupt/recurrent   \\ 
  &  & SynM-GrdRec  &   20,480    &  784       &   192 (0.94$\%$)  & Gradual/recurrent \\ 
 &   & SynF-AbrRec      &   20,480   &  784       &    193 (0.94$\%$) & Abrupt/recurrent   \\
  &  & SynF-GrdRec     &   20,480   &   784    &    204 (1.00$\%$) & Gradual/recurrent  \\ 
    \bottomrule[1.8pt]
    \end{tabular}
    }
    \vspace{-5mm}
\end{table}

\begin{table*}[ht]
    \small
    \centering
    \renewcommand{\arraystretch}{1} 
    \caption{\pamirevise{Performance comparison under unknown drifts in a discrete setting with higher scores indicating superior performance.
    We highlight the \textbf{\underline{best}}  and the \textbf{second best} result in each row.}}\vspace{-2mm}
    \label{tab:overall_resutls_un_dis}
    \resizebox{2.1\columnwidth}{!}{
        \begin{tabular}{ c c c c c c c c c c c c c c c c}
    
    \toprule[1.5pt]
    \multirow{2}{*}{Model Class} & \multirow{2}{*}{Model} & \multicolumn{2}{c}{Ion.}  &   \multicolumn{2}{c}{Pima}  &   \multicolumn{2}{c}{Satellite} & \multicolumn{2}{c}{Mamm.} & \multicolumn{2}{c}{\revise{BGL}}  & \multicolumn{2}{c}{NSL-KDD}  &   \multicolumn{2}{c}{KDD99} \\ 
    &   &   AUCROC   &  AUCPR  & AUCROC   &  AUCPR &  AUCROC   &  AUCPR  & AUCROC   &  AUCPR  & AUCROC   &  AUCPR & AUCROC   &  AUCPR & AUCROC   &  AUCPR \\
    
    \midrule[0.5pt]
  \multirow{4}{*}{Traditional} &  LOF~\cite{breunig2000lof}  & 
0.874 & 0.827  & 0.542 & 0.371 & 0.598 & 0.481 & 0.720 & 0.089 & 0.542 & 0.206 & 0.586 & 0.428 & 0.653 & 0.359 \\
  &  IF~\cite{liu2008isolation}  &  0.860 & 0.817 &  0.677 &  0.502 &  0.676 & 0.375 &  0.867  & 0.211 &  0.823 & 0.295 & 0.530 & 0.577 & 0.784 & 0.406 \\
   &  KNN  &  0.929 & 0.932 &0.615 & 0.457 & 0.677 & 0.539 & 0.839 & 0.156 & 0.765 & 0.274 & 0.897 &  0.899 & 0.946 & 0.902 \\
  &  STORM~\cite{angiulli2007detecting} & 0.640 & 0.526 & 0.529 & 0.373 & 0.680 & 0.452 & 0.615 & 0.418 & 0.203 & 0.043 & 0.513 &  0.138 & 0.913 &  0.822 \\

    \midrule[0.5pt]
    \multirow{3}{*}{Incremental} &  RRCF~\cite{guha2016robust}  & 0.586 & 0.411 & 0.575 & 0.393 & 0.553 & 0.356 & 0.713 & 0.524 & 0.540 & 0.076 & 0.604 & 0.534 & 0.773 & 0.347 \\
    &   MStream~\cite{bhatia2021mstream}  & 0.681 & 0.486 & 0.524 & 0.440 & 0.647 & 0.457 & 0.798 &	0.076 & 0.531 & 0.105 & 0.759 & 0.716 & 0.958 & 0.912 \\
         
    &   MemStream~\cite{bhatia2022memstream}    &   0.821  &  0.672 &   0.703  & 0.551   &  0.722 & 0.682 & 0.902 & 0.225 & 0.694 & 0.144 & \textbf{0.988}  &   \textbf{\underline{0.967}} & 0.979 &   0.857  \\

    \midrule[0.5pt]
    \multirow{8}{*}{Ensemble} &   HS-Trees~\cite{tan2011fast}  & 0.687 & 0.574 & 0.667 & 0.344 & 0.512 & 0.348 & 0.797 & 0.623 & 0.599 & 0.174 & 0.806 & 0.735 & 0.901 & 0.728 \\
    &   iForestASD~\cite{ding2013anomaly}  &  0.744 & 0.601 & 0.515 & 0.356 & 0.642 & 0.451 & 0.575 & 0.031 & 0.701 & \textbf{0.382} &  0.511 & 0.483 & 0.532 & 0.227 \\
   &   RS-Hash~\cite{sathe2016subspace}    & 0.743 & 0.502 & 0.518 & 0.372 & 0.640 & 0.586 & 0.776 & 0.622 & 0.436 & 0.245 & 0.684   & 0.524 &  0.783 & 0.707 \\
    &   LODA~\cite{pevny2016loda}   & 0.514 & 0.373 & 0.501 & 0.347 & 0.500 & 0.316 & 0.500 & 0.023 & 0.523 & 0.074 & 0.504 & 0.535  & 0.507 & 0.197 \\    
    &   Kitsune~\cite{mirsky2018kitsune}   & 0.920 & 0.896 & 0.590 & 0.451 & 0.732 & 0.673 & 0.603 & 0.202 &	0.514 & 0.074 & 0.947 & 0.918 & 0.982 & \textbf{\underline{0.993}} \\ 
    &   xStream~\cite{manzoor2018xstream}  & 0.773 & 0.591 & 0.656 & 0.583 & 0.659 & 0.533 & 0.847 & \textbf{\underline{0.630}} & 0.623 & 0.356 & 0.540 & 0.327 & 0.954 & 0.881 \\
    &  PIDForest~\cite{gopalan2019pidforest}   & 0.821 & 0.718 & 0.669 & 0.474 & 0.718 & 0.543 & 0.847 & 0.202 & 0.791 & 0.300 & 0.503 & 0.561 & 0.864 & 0.772 \\
    &   ARCUS~\cite{yoon2022adaptive}    & 0.919 & 0.894 & 0.607 & 0.420 & \textbf{0.797} & 0.560 & 0.812 & 0.261 & 0.768 & 0.185 & 0.262 & 0.365 & 0.972 & 0.807 \\
    
  \midrule[0.5pt]
  
     \multirow{4}{*}{\pamirevise{Drift-adaptive}}  &   \pamirevise{D$^3$R~\cite{wang2023drift}} & \pamirevise{0.758} & \pamirevise{0.605}& \pamirevise{0.546} & \pamirevise{0.363} &  \pamirevise{0.731} & \pamirevise{0.577} & \pamirevise{0.748}& \pamirevise{0.112} &\pamirevise{0.872} & \pamirevise{0.436}& \pamirevise{0.910} & \pamirevise{0.817} & \pamirevise{0.947} & \pamirevise{0.758}\\
     
     &  \pamirevise{SARAD~\cite{dai2024sarad}} &\pamirevise{0.926} & \pamirevise{0.897}& \pamirevise{0.645} & \pamirevise{0.578} &  \pamirevise{0.767} & \pamirevise{0.608} & \pamirevise{\textbf{\underline{0.938}}}& \pamirevise{0.204} &\pamirevise{0.754} & \pamirevise{0.103}& \pamirevise{0.891} & \pamirevise{0.654} & \pamirevise{\textbf{0.987}} & \pamirevise{0.916}\\

     &  METER~\cite{zhu2023meter}  &  \textbf{0.950}  &  \textbf{0.956}  &     \textbf{0.733}  &  \textbf{0.654} & 0.796 &  \textbf{\underline{0.777}}  &   \textbf{0.913}  &  0.491  &  \textbf{0.895} & 0.369 & 0.982 & 0.963  & 0.973  &  0.853 \\
 &   \name  &  \textbf{\underline{0.972}}  &  \textbf{\underline{0.961}}  &  \textbf{\underline{0.812}}  &  \textbf{\underline{0.686}}  & \textbf{\underline{0.828}}  &  \textbf{0.762} &   0.910  &  \textbf{0.617}  &  \textbf{\underline{0.906}} & \textbf{\underline{0.615}} & \textbf{\underline{0.991}}  &  \textbf{0.965}  & \textbf{\underline{0.990}}  &  \textbf{0.931}  \\
    
    \bottomrule[1.5pt]
        \end{tabular}
    } \vspace{-4mm}
\end{table*}

\vspace{-6mm}
\subsection{Baseline Methods}
\label{sec:baseline}
\pamirevise{In our evaluation, we compare \name with 18 baseline methods in four categories:}
(1) Representative anomaly detection algorithms, including Local Outlier Factor (LOF)~\cite{breunig2000lof}, Isolation Forest (IF)~\cite{liu2008isolation}, k-Nearest Neighbors (KNN), and STORM~\cite{angiulli2007detecting}.
(2) \pamirevise{Incremental methods}, including RRCF~\cite{guha2016robust}, 
MStream~\cite{bhatia2021mstream}, and MemStream~\cite{bhatia2022memstream}.
(3) \pamirevise{Ensemble methods}, comprising HS-Trees~\cite{tan2011fast}, iForestASD~\cite{ding2013anomaly}, RS-Hash~\cite{sathe2016subspace}, LODA~\cite{pevny2016loda}, Kitsune~\cite{mirsky2018kitsune}, xStream~\cite{manzoor2018xstream}, PIDForest~\cite{gopalan2019pidforest}, and ARCUS~\cite{yoon2022adaptive}.
\pamirevise{
(4) Advanced drift-adaptive architectures, namely D$^3$R~\cite{wang2023drift}, SARAD~\cite{dai2024sarad}, and our earlier work METER~\cite{zhu2023meter}.}
\pamirevises{Overall, the baselines span diverse underlying model families, including three proximity-based (distance/density) methods, four projection-based (hashing/random projection) methods, five tree-based methods, and six neural network–based methods, as introduced below.}

\pamirevises{
\begin{itemize}[leftmargin=*]
\item \powerpoint{LOF}~\cite{breunig2000lof} is a density-based method that measures the local deviation of a point with respect to its neighbors. 
\item\powerpoint{IF}~\cite{liu2008isolation} is a tree-based method that isolates outliers via recursive data partitioning using decision trees.
\item \powerpoint{KNN} is a distance-based method that identifies outliers as points with few neighbors within a specified distance. 
\item \powerpoint{STORM} ~\cite{angiulli2007detecting} is a distance-based method that employs a sliding window approach to detect deviations from the window's average behavior. 
\item \powerpoint{RRCF} ~\cite{guha2016robust} is a tree-based method that detects anomalies using a forest of random cut trees.
\item \powerpoint{MStream}~\cite{bhatia2021mstream} is a hashing-based method that uses locality-sensitive hashing to identify unusual group anomalies. 
\item \powerpoint{MemStream}~\cite{bhatia2022memstream} is a neural network-based method that models streaming trends using a denoising autoencoder with a First-In-First-Out (FIFO) memory. 
\item \powerpoint{HS-Trees}~\cite{tan2011fast} is a tree-based method that quickly builds its tree structure based on the dimensionality of the data space.
\item \powerpoint{iForestASD}~\cite{ding2013anomaly} is a tree-based method that integrates IF with sliding windows for streaming anomaly detection.
\item \powerpoint{RS-Hash}~\cite{sathe2016subspace} is a hashing-based method that detects anomalies using an ensemble of randomized subspace hashing and frequency-based scores.
\item \powerpoint{LODA} ~\cite{pevny2016loda} is a random projection–based method that employs an ensemble of one-dimensional histogram detectors for anomaly detection.
\item \powerpoint{Kitsune} ~\cite{mirsky2018kitsune} is a neural network–based method that detects anomalies using an ensemble of autoencoders. 
\item \powerpoint{xStream} ~\cite{manzoor2018xstream} is a random projection–based method that detects anomalies via density estimation on low-dimensional projections.
\item \powerpoint{PIDForest}~\cite{gopalan2019pidforest} is a tree-based method that trains each tree on a feature subset selected via partial identification.
\item \powerpoint{ARCUS}~\cite{yoon2022adaptive} is a neural network–based method that employs a model pool to detect anomalies and dynamically updates its members under concept drift.
\item
\powerpoint{D$^3$R}~\cite{wang2023drift} is a neural network–based method that detects anomalies via dynamic decomposition and diffusion-based reconstruction.
\item
\powerpoint{SARAD}~\cite{dai2024sarad} is a neural network–based method that extracts spatial association descent via a Transformer-based autoencoder to detect anomalies.
\item \powerpoint{METER}~\cite{zhu2023meter} is a neural network–based method that detects drift via evidential deep learning and adapts the detector with a static predefined decision threshold.
\end{itemize}
}

\vspace{-3mm}
\subsection{Evaluation Metrics} \label{sec:metrics}
We utilize AUCROC and AUCPR metrics for evaluation.
AUCROC calculates the area under the receiver operating characteristic (ROC) curve, plotting the false positive rate (FPR) against the true positive rate (TPR) across various thresholds.
AUCPR measures the area under the precision-recall (PR) curve, displaying the relationship between precision and recall at differing thresholds.
Both metrics range from $0$ to $1$, with higher values indicating superior performance.


\vspace{-3mm}
\subsection{\pamirevises{Implementation Details}}
\label{sec:implementation_details}
\pamirevises{The encoder and decoder of \name are implemented as symmetrically structured DNNs with 2–10 layers.
The latent dimension of the autoencoder is set to the number of principal components to ensure a minimum of 70\%  explained variance following prior work~\cite{yoon2022adaptive,kim2020rapp}.
The IEC is implemented as a two-layer DNN with ReLU activations.
The pseudo-labeling threshold $\mu_{p}$ is selected from $[0.05,0.5]$ with a step size of 0.05, while the concept uncertainty threshold $\mu_{e}$ is searched within $[0.005,0.4]$.
The focal exponent $\gamma$ in $\mathcal{L}_{IEC}$ is fixed to 2, and the uncertainty calibration weight $\lambda$ in the anomaly score is set to 0.6.
The offline update threshold $\mu_{o}$ is in the range of 0.1 to 1 of the window size, and the sizes of the sliding windows $\mathcal{W}_N$ and $\mathcal{W}_C$ are each set at 64.
Hyperparameters $\tau$ and $\kappa$ are fixed at $0.95$ and $0.8$, respectively.
The historical data ratio $h_r$ is set to 0.2.
\name is trained using the Adam optimizer with an initial learning rate of $1e$-2, decayed exponentially by a factor of 0.96 over 2000 epochs.
D$^3$R and SARAD are implemented using the official source code released by the respective authors\footnote{D$^3$R: \url{https://github.com/ForestsKing/D3R}, SARAD: \url{https://github.com/daidahao/SARAD/tree/main}}.
All other baseline implementations follow the settings in our prior work~\cite{zhu2023meter}.
We develop a data generator to simulate the generation of streaming data and report the mean performance over five independent runs.
The full implementation and experimental scripts are available at \url{https://github.com/zjiaqi725/DyMETER}.
}

\ignore{
We implement LOF and IF using the \emph{scikit-learn} library~\cite{pedregosa2011scikit}, and KNN using the pyod library~\cite{zhao2019pyod}.
The open-source PySAD library~\cite{yilmaz2020pysad} is used to implement STORM~\cite{angiulli2007detecting}, RRCF~\cite{guha2016robust}, HS-Trees~\cite{tan2011fast}, iForestASD~\cite{ding2013anomaly}, RS-Hash~\cite{sathe2016subspace}, LODA~\cite{pevny2016loda}, and xStream~\cite{manzoor2018xstream} with default parameters.
For other baseline methods such as MStream~\cite{bhatia2021mstream}, MemStream~\cite{bhatia2022memstream}, Kitsune~\cite{mirsky2018kitsune}, PIDForest~\cite{gopalan2019pidforest}, ARCUS~\cite{yoon2022adaptive}, we adopt the official implementations, using the recommended parameter settings.
For ARCUS, we use the base model RAPP~\cite{kim2020rapp}.
In cases where default parameter values are not provided, we conduct a grid search to find the optimal parameters that yield the best performance.
Adam~\cite{adam} is used as an optimizer in all learning-based anomaly detection models with a learning rate searched in $0.1 \unsim 1e$-3.
}

\begin{table*}[ht]
    \small
    \centering
    \renewcommand{\arraystretch}{1} 
    \caption{\pamirevise{Performance comparison of unknown drifts in a continuous setting.}}
    \vspace{-2mm}
    \label{tab:overall_resutls_un_con}
    
    \resizebox{2.1\columnwidth}{!}{
        \begin{tabular}{ c c c c c c c c c c c c c c c c c c }
    
    \toprule[1.5pt]
    \multirow{2}{*}{Model Class} & \multirow{2}{*}{Model} & \multicolumn{2}{c}{\violet{A.R.}}  &   \multicolumn{2}{c}{\violet{I.B.}}  &   \multicolumn{2}{c}{\violet{NASA}}  &   \multicolumn{2}{c}{\violet{GaitPhase}} &\multicolumn{2}{c}{\violet{EPG}} &\multicolumn{2}{c}{\violet{ECG}} &  \multicolumn{2}{c}{M.T.}  &
    \multicolumn{2}{c}{CPU.} \\
    &   &   AUCROC   &  AUCPR  &  AUCROC   &  AUCPR &  AUCROC   &  AUCPR  & AUCROC   &  AUCPR  & \blue{AUCROC} & \blue{AUCPR}& \blue{AUCROC} &  \blue{AUCPR}  & \blue{AUCROC} &  \blue{AUCPR} & \blue{AUCROC} &  \blue{AUCPR} \\
    
    \midrule[0.5pt]
  \multirow{4}{*}{Traditional} &  LOF~\cite{breunig2000lof}  & 0.615 & 0.170 &  0.782 & 0.180 &  0.653 &  0.477 & 0.821 &  0.328 & 0.934 &\blue{0.679} & \blue{0.670} &\blue{0.016}  &  0.501 & 0.141 & 0.560 &  0.112 \\

  &  IF~\cite{liu2008isolation}  & 0.938 & 0.568 & 0.850 & 0.012 & 0.766 & 0.018 & 0.747  & 0.013 & \blue{0.811} & \blue{0.552}&  \blue{0.668} & \blue{0.005} &  0.829 & 0.573  & 0.817  & \textbf{\underline{0.760}} \\

   &  KNN  & 0.508 & 0.132 &  0.540 & 0.165  &  0.475 & 0.478 & 0.532 & 0.200 & \blue{0.083} & \blue{0.001}&  \blue{0.247} &\blue{0.002}& 0.759 & 0.255 &0.724 & 0.452 \\
   
  & STORM~\cite{angiulli2007detecting}  & 0.597 & 0.313 & 0.534 &  0.072 & 0.593 & 0.041 & 0.602 & 0.137 & \blue{0.578} &\blue{0.436} & \blue{0.662}&\blue{\textbf{0.523}}& 0.604 & 0.127 & 0.667 & 0.605 \\
  
    \midrule[0.5pt]
    \multirow{3}{*}{Incremental} &   RRCF~\cite{guha2016robust}  &  0.834 & 0.175 & 0.503 & 0.079 & 0.515 & 0.161 & 0.586 &  0.118 & \blue{0.814} & \blue{0.498} & \blue{0.387} &\blue{0.002} & 0.628 & 0.153 &  0.617 &  0.368 \\
    &   MStream~\cite{bhatia2021mstream}   & 0.882 &	0.556 & 0.672 &	0.389 & 0.673 & 0.328 & 0.816 & 0.326 &	\blue{0.824} &\blue{0.621} & \blue{0.721}& \blue{0.284} & 0.860 &	0.505 & 0.794 & 0.443 \\
    
    &   MemStream~\cite{bhatia2022memstream}  &  \textbf{0.984} & 0.354 & 0.851 & 0.016 & 0.468 & 0.058 &  \textbf{0.826}   &  0.120   & \blue{0.930} & \textbf{0.656} & \blue{\textbf{0.780}} & \blue{0.007} &  0.825  &  0.573 & 0.831  &  0.227  \\
    
    \midrule[0.5pt]
    \multirow{8}{*}{Ensemble} &   HS-Trees~\cite{tan2011fast}  & 0.585 & 0.176 & 0.487 & 0.071 & 0.591 &  0.172 & 0.624 & 0.189 &\blue{0.531}&\blue{0.334}& \blue{0.621}&\blue{0.439}&  0.617 & 0.359 & 0.678 & 0.585 \\
    
    &   iForestASD~\cite{ding2013anomaly}   & 0.517 & 0.202 & 0.641 & 0.137 & 0.511 & 0.086 & 0.689 & 0.208 & \blue{0.782} &\blue{0.470} & \blue{0.733}& \blue{0.006}& 0.738 & 0.231 & 0.755 & 0.153 \\
    
    &   RS-Hash~\cite{sathe2016subspace}   & 0.567 & 0.220 & 0.498 & 0.074 & 0.496 & 0.017 & 0.614 &  0.208 & \blue{0.552}& \blue{0.186}& \blue{0.584} &\blue{0.203} & 0.607 & 0.549 & 0.712 &  0.467 \\
    
    &   LODA~\cite{pevny2016loda}  & 0.645  & 0.177 & 0.553 & 0.092 &  0.516 & 0.128 & 0.502 &  0.096 & \blue{0.595} & \blue{0.182} &\blue{0.721} &\blue{0.077} & 0.503  & 0.100 & 0.500 &  0.083 \\

    &   Kitsune~\cite{mirsky2018kitsune}   & 0.517 & 0.154 & 0.508 & 0.072 & 0.512 & 0.172 & 0.607 & 0.249 & \blue{0.897} &\blue{0.020}&  \blue{0.726}&\blue{0.231}& 0.684 & 0.416 &0.824 & 0.669 \\ 
   &   xStream~\cite{manzoor2018xstream}   & 0.802  &  0.153  & 0.493  & 0.072  &  0.512  &  0.029 &  0.623 &  0.187  & \blue{0.687} &\blue{0.158} & \blue{0.705}& \blue{0.365} & 0.696 &  0.596 & 0.730 &  0.195 \\
   
    &  PIDForest~\cite{gopalan2019pidforest}  &  0.924 & 0.560 & 0.829 & 0.300 & 0.641 & 0.012 &  0.805 & 0.055 & \blue{0.836}&\blue{0.238}& \blue{0.683}&\blue{0.006}&  0.789 & 0.389 & 0.881 & 0.439 \\

    & ARCUS~\cite{yoon2022adaptive}   &  0.547 & 0.114 & 0.742 & 0.543 & 0.768 & 0.449 & 0.815 & \textbf{0.366} & \blue{0.885} &\blue{\textbf{\underline{0.724}}} & \blue{0.682}&\blue{0.340} &   0.376 & 0.511 & 0.678 & 0.128 \\
  \midrule[0.5pt] 

    \multirow{4}{*}{\pamirevise{Drift-adaptive }}  &   \pamirevise{D$^3$R~\cite{wang2023drift}} & \pamirevise{0.942} & \pamirevise{0.355} &\pamirevise{0.822} & \pamirevise{0.040}& \pamirevise{0.623} & \pamirevise{0.209} & \pamirevise{\textbf{\underline{0.879}}} & \pamirevise{0.031}& \pamirevise{0.906} & \pamirevise{0.220} & \pamirevise{0.726} & \pamirevise{0.285} & \pamirevise{0.627} & \pamirevise{0.197} & \pamirevise{0.719} & \pamirevise{0.301}\\
    
     &  \pamirevise{SARAD~\cite{dai2024sarad}} &\pamirevise{0.926} & \pamirevise{0.584} & \pamirevise{\textbf{0.889}} &\pamirevise{0.392} & \pamirevise{0.712}& \pamirevise{0.354}& \pamirevise{0.821}& \pamirevise{0.285}& \pamirevise{0.934}& \pamirevise{0.521}& \pamirevise{0.686} & \pamirevise{0.135}& \pamirevise{\textbf{\underline{0.871}}} & \pamirevise{\textbf{\underline{0.665}}}& \pamirevise{0.847} & \pamirevise{0.692} \\

    &   METER~\cite{zhu2023meter}  & \textbf{\underline{0.998}} &  \textbf{0.798}  &  0.874 & \textbf{0.545}  & \textbf{0.811}  & \textbf{0.489} & 0.824 & 0.335 & \blue{\textbf{\underline{0.968}}}& \blue{0.625}& \blue{0.778}& 0.495 & 0.842 &  \textbf{0.652}  & \textbf{0.908} & 0.715\\
  &  \name  & \textbf{\underline{0.998}} &  \textbf{\underline{0.856}}  & \textbf{\underline{0.896}}  &  \textbf{\underline{0.612}}  & \textbf{\underline{0.854}}  & \textbf{\underline{0.527}} & 
  \textbf{0.832} & \textbf{\underline{0.384}}&   \textbf{0.956}& \blue{0.645}& \textbf{\underline{0.810}}& \textbf{\underline{0.568}}& \textbf{0.866} &  0.629  & 
  
  \textbf{\underline{0.911}} & \textbf{0.723}\\
  
    \bottomrule[1.5pt]
        \end{tabular}
    }
    \vspace{-2mm}
\end{table*}

\noindent \textbf{Experimental Environment.}
\blue{All the experiments are conducted in a server with Xeon(R) Silver 4114 CPU @ 2.2GHz (10 cores), 256G memory, and GeForce RTX 2080 Ti.
All the models are implemented in PyTorch 1.10.0 with CUDA 10.2.}


\begin{table}[t]
    \small
    \centering
    \renewcommand{\arraystretch}{0.85}
    \caption{\pamirevise{Performance comparison of known drifts in a continuous setting in AUCROC metric.}}
    \vspace{-2mm}
\label{tab:overall_resutls_kn_con}
    \resizebox{1\columnwidth}{!}{
        \begin{tabular}{ c c c c c c c}
    
    \toprule[1.5pt]
    \multirow{2}{*}{Model Class} & \multirow{2}{*}{Model} & INSECTS  &   INSECTS  & INSECTS &  INSECTS  & \multirow{2}{*}{Time (s)} \\
    & & -Abr& -Inc& -IncGrd  &-IncRec &  \\
    \midrule[0.5pt]
  \multirow{4}{*}{Traditional} &  LOF~\cite{breunig2000lof}  & 0.578 & 0.556 &  0.589 & 0.526 & 180 \\
    &  IF~\cite{liu2008isolation}  &  0.679 & 0.632 & 0.697 & 0.593  & 67 \\
    &  KNN  &  0.666 & 0.597 &  0.673 & 0.553 & 105 \\
&  STORM~\cite{angiulli2007detecting} & 0.408 & 0.441 & 0.446 & 0.449 & 122\\

    \midrule[0.5pt]
    \multirow{3}{*}{Incremental} &   RRCF~\cite{guha2016robust} & 0.600 & 0.579 &0.624 & 0.593 &  121 \\
    &   MStream~\cite{bhatia2021mstream}  & 0.703 & 0.698 &	\textbf{\underline{0.788}} & 0.672 & 18\\
    &   MemStream~\cite{bhatia2022memstream}  &  0.753  &  0.348 &   0.728 &  0.361  & 109 \\
    
    \midrule[0.5pt]
    \multirow{8}{*}{Ensemble} &   HS-Trees~\cite{tan2011fast}  & 0.499 & 0.507 & 0.497 & 0.499  & 302\\
    &   iForestASD~\cite{ding2013anomaly} & 0.599 & 0.589 & 0.616 & 0.575 & 7985\\
    &   RS-Hash~\cite{sathe2016subspace}   & 0.484 & 0.509  & 0.459 & 0.506 &225\\
    &   LODA~\cite{pevny2016loda}  &  0.498 & 0.503 &0.496 & 0.499 & 831\\
    &   Kitsune~\cite{mirsky2018kitsune}   &0.759 & 0.584 & 0.730 & 0.594 & 164 \\ 
    &   xStream~\cite{manzoor2018xstream}  &  0.514 & 0.516 & 0.533 & 0.504 & 408\\
    &  PIDForest~\cite{gopalan2019pidforest}  & 0.757 & 0.675 & 0.748 & 0.631 & 18047 \\
    &   ARCUS~\cite{yoon2022adaptive}  &  0.601 &   0.597  & 0.576 &   0.632  & 79\\
    
  \midrule[0.5pt]
  
 \multirow{4}{*}{\pamirevise{Drift-adaptive}}  &   \pamirevise{D$^3$R~\cite{wang2023drift}} & \pamirevise{0.776} & \pamirevise{0.786} &\pamirevise{0.713} & \pamirevise{0.743}& \pamirevise{486} \\
 
     &  \pamirevise{SARAD~\cite{dai2024sarad}} & \pamirevise{0.731} & \pamirevise{0.739} &\pamirevise{0.651} & \pamirevise{0.728}& \pamirevise{324}\\
 
 & METER~\cite{zhu2023meter}  &   \textbf{0.816} &  \textbf{0.795}  & 0.712 &  \textbf{0.794} & 88  \\
  & \name  & \textbf{\underline{0.832}} &  \textbf{\underline{0.814}}  & \textbf{0.756}
  &  \textbf{\underline{0.842}} & 136 \\
    \bottomrule[1.5pt]
        \end{tabular}
    }
     \vspace{-3mm}
\end{table}

\subsection{Effectiveness}
\label{sec:exp_effectiveness}
We comprehensively evaluate \name across a diverse array of datasets, comparing it against various state-of-the-art methods, with results detailed in Tables~\ref{tab:overall_resutls_un_dis} to \ref{tab:overall_resutls_kn_dis}.
In these experiments, \name consistently outperforms competitors across different experimental conditions, data dimensions, and types of drift, highlighting its exceptional effectiveness and robustness. Notably, \name exhibits significant improvements in performance over the initial version of METER in nearly all scenarios.
A detailed performance analysis follows.

\noindent
\powerpoint{\revise{Comparison on real-world datasets.}}
We evaluate \name against various baselines on 19 real-world datasets, with results detailed in Tables~\ref{tab:overall_resutls_un_dis}, \ref{tab:overall_resutls_un_con}, and \ref{tab:overall_resutls_kn_con}, where \name consistently excel across different drift types and settings.
\pamirevise{For instance, among the 14 results displayed in Table~\ref{tab:overall_resutls_un_dis}, our method achieves the highest performance in 9 cases and the second highest in the remaining 4 cases among all four classes of baselines}.
By adopting the shingling technique for time series preprocessing as~\cite{gopalan2019pidforest,zhu2023meter}, \name effectively captures temporal dependencies, enabling \name to deliver \pamirevise{the best performance on 6 out of the 8 datasets presented in Table~\ref{tab:overall_resutls_un_con}.}
In the continuous setting featuring various known drifts, such as abrupt and incremental, \name consistently demonstrates superior performance, achieving an average improvement of 4.1\% over the next suboptimal method METER.
\pamirevise{
Although methods such as D$^3$R and SARAD employ architectural designs to handle concept drift without relying on incremental updates or ensembles, and SARAD demonstrates notable performance on certain continuous datasets (e.g., M.T. and I.B.), their effectiveness is less consistent across diverse benchmarks.
}
Furthermore, \name shows remarkable computational efficiency, with inference time in Table~\ref{tab:overall_resutls_kn_con}
lower than half of the baseline methods, while providing superior performance.

\noindent
\powerpoint{\pamirevise{Comparison on synthetic datasets.}}
We further evaluate \name on four synthetic datasets, with the outcomes detailed in Table \ref{tab:overall_resutls_kn_dis}.
\pamirevise{
Despite the challenges posed by increased data dimensions, \name consistently surpasses other baseline methods, including recent advanced drift-adaptive architectures such as D$^3$R and SARAD, and shows marked improvements over our initial version METER.}
Moreover, although ARCUS occasionally reached or slightly exceeded \name in the gradual drift scenarios, its performance lacks consistency.
Overall, \name's consistently high performance in abrupt and gradual drift scenarios, combined with reasonable computational demands, establishes it as a potent solution for OAD in dynamic environments.

\pamirevise{
\noindent
\powerpoint{Effectiveness of online and offline adaptation.}
To further substantiate the effectiveness of online adaptation and offline fine-tuning, we partition the INSECTS-Abr dataset into five sequential and equal subsets, denoted as $p_1$ to $p_5$.
Then, we further split each subset into one training set and one test set, and derive $p_{1-train}/p_{1-test}$ to $p_{5-train}/p_{5-test}$ correspondingly.
Using these subsets, we conduct two groups of experiments.
In Group I, we train \name on $p_{1-train}$ and then fine-tune \name on $p_{2-train}$ to $p_{5-train}$ respectively, and report the detection performance of \name.
As for the experiments of Group II, we only train \name on $p_{1-train}$ once, and then report the trained \name on all test sets, namely $p_{1-test}$ to $p_{5-test}$.
Results summarized in Table~\ref{tab:reccurring} show that (i) offline adaptation (Group I) can even degrade performance compared to the model trained once with only online adaptation, as observed on $p_{4-test}$ and $p_{5-test}$, and (ii) the model with online adaptation (Group II) achieves an average AUCROC of 0.795, which is comparable to the average performance of Group I with costly offline fine-tuning and only slightly lower than 0.811 obtained on $p_{1-test}$.
These findings suggest that offline adaptation is not always effective for OAD tasks, which may be attributed to the information bottleneck inherent in reconstruction-based models:
if it is too restrictive, the model \textit{underfits} and fails to reconstruct even normal data, whereas if it is too loose, the model \textit{overfits} and reconstructs anomalies successfully.
}

\begin{table}[t]
    \small
    \centering
    \renewcommand{\arraystretch}{0.7}
    \caption{\pamirevise{Performance comparison of known drifts in a discrete setting in AUCROC metric.}} 
    \vspace{-2mm}
    \label{tab:overall_resutls_kn_dis}
    
    \resizebox{1\columnwidth}{!}{
        \begin{tabular}{ c c c c c c c}
    
    \toprule[1.5pt]
    \multirow{2}{*}{Model Class} & \multirow{2}{*}{Model} & SynM  &  SynM  &   SynF &  SynF  & \multirow{2}{*}{Time (s)} \\
  &  & -AbrRec & -GrdRec & -AbrRec & -GrdRec & \\
  \midrule[0.5pt]
   \multirow{4}{*}{Traditional} &  LOF~\cite{breunig2000lof}  & 0.554 & 0.539 & 0.473 & 0.503 & 620\\
   &  IF~\cite{liu2008isolation}  &  0.565 & 0.544 & 0.497 & 0.478 & 328\\
   &  KNN  &  0.550 & 0.584 & 0.515 & 0.513 & 55 \\
  &  STORM~\cite{angiulli2007detecting} &  0.513 & 0.516 & 0.521 & 0.530 & 323\\   
    \midrule[0.5pt]
    \multirow{3}{*}{Incremental} &   RRCF~\cite{guha2016robust} & 0.695 & 0.666 & 0.681 & 0.715 & 142\\
    &  MStream~\cite{bhatia2021mstream}  & 0.473 & 0.608 & 0.623 & 0.507 & 31\\
    &  MemStream~\cite{bhatia2022memstream}  &  0.517  &  0.496 & 0.523  &   0.472 &  549 \\
    \midrule[0.5pt]
    \multirow{8}{*}{Ensemble} &   HS-Trees~\cite{tan2011fast} & 0.497 & 0.503 & 0.500 & 0.511 & 139 \\
    &   iForestASD~\cite{ding2013anomaly} &  0.523 & 0.514 & 0.533 &  0.506 & 2580 \\
   &   RS-Hash~\cite{sathe2016subspace}   & 0.482 & 0.506 & 0.470 & 0.506 & 102 \\
    &   LODA~\cite{pevny2016loda}  & 0.504 & 0.500 & 0.502 & 0.500 & 328\\
    &   Kitsune~\cite{mirsky2018kitsune}   & 0.544  & 0.502 & 0.570  & 0.523 & 323 \\ 
    &   xStream~\cite{manzoor2018xstream}  & 0.641 & 0.658  & 0.625 & 0.605 & 997\\
    &  PIDForest~\cite{gopalan2019pidforest}  &  0.514 & 0.546 &  0.513 &  0.503  &  8176 \\
    &   ARCUS~\cite{yoon2022adaptive}   & \textbf{0.901} & \textbf{\underline{0.894}} & 0.759 & \textbf{\underline{0.787}} & 120 \\
    
  \midrule[0.5pt]
  
  \multirow{4}{*}{\pamirevise{Drift-adaptive}}  &  \pamirevise{D$^3$R~\cite{wang2023drift}} & \pamirevise{0.761} & \pamirevise{0.699} &\pamirevise{0.672} & \pamirevise{0.619}& \pamirevise{508}\\
 
 &  \pamirevise{SARAD~\cite{dai2024sarad}} & \pamirevise{0.693} & \pamirevise{0.633} &\pamirevise{0.757} & \pamirevise{0.670}& \pamirevise{368}\\

 &   METER~\cite{zhu2023meter}  &   0.879 & 0.825 & \textbf{0.806} & 0.632 & 72 \\
  &  \name  &   \textbf{\underline{0.906}} &\textbf{0.886} & \textbf{\underline{0.858}} & \textbf{0.728} & 142 \\
    \bottomrule[1.5pt]
        \end{tabular}
    } 
    \vspace{-3mm}
\end{table}

\subsection{\pamirevises{Ablation Study}}
\label{sec:exp_ablation}

\pamirevises{In this section, we conduct ablation studies to evaluate both the effectiveness of individual components and the contributions of their dedicated mechanism designs under diverse and realistic drift scenarios, as detailed below.}

\begin{figure}[t]
    \centering
\includegraphics[width=1\linewidth]{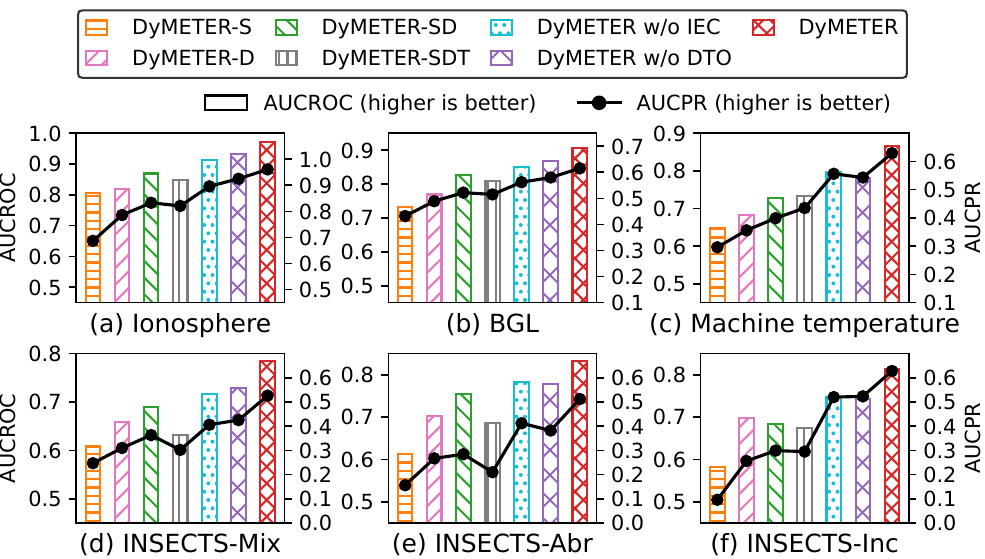}
\caption{\pamirevises{Ablation analysis on module effectiveness via AUCROC and AUCPR for \name and its six variants.
    }}
    \label{fig:ablation_auc}
    \vspace{-4mm}
\end{figure}

\noindent
\textbf{Module effectiveness.}
\pamirevises{We evaluate the contribution of each \name module on six real-world datasets subject to diverse and often unknown concept drifts.}
\pamirevise{We define six \name variants:
\name-S and \name-D, which include only the SCD or DSD module, respectively;
\name-SD and \name-SDT retain SCD with DSD or DTO, respectively, while excluding all other modules;
and \name w/o IEC and \name w/o DTO, which respectively omit the IEC or DTO module from \name.}
\pamirevises{
The results in Figure~\ref{fig:ablation_auc} consistently confirm that each module contributes considerably to enhancing detection performance under concept drift.
Notably, the results indicate that DSD plays a critical role, with its removal leading to up to an 18.1\% drop in AUCROC across the evaluated datasets, and its integration with SCD, which provides a stable basis for adaptation, further improves overall effectiveness.
In addition, both IEC and DTO are important for effective online anomaly detection.
IEC identifies drift-affected instances to trigger adaptation only when necessary, while DTO continuously recalibrates the decision boundary as concepts evolve and yields a 7.1\% AUCROC gain across the evaluated datasets.}

\begin{table}[t]
    \small
    \centering
\renewcommand{\arraystretch}{1.2}
\caption{\pamirevise{Performance comparison of \name with offline and online adaptation measured in AUCROC.}}
    \vspace{-2mm}
    \label{tab:reccurring}
    \resizebox{1\columnwidth}{!}{
        \begin{tabular}{ c c c c c c c }
    \toprule[1.5pt]
   	Variant & $p_{1-test}$ & $p_{2-test}$ & $p_{3-test}$ & $p_{4-test}$ & $p_{5-test}$ & Average \\ 
    \hline	
        Group I & 0.811 & 0.802 & 0.780 & 0.792 & 0.795 & 0.796 $\pm$ 0.010 \\
	Group II & 0.811 & 0.783 & 0.766 & 0.812 & 0.802 & 0.795 $\pm$ 0.018\\
    \bottomrule[1.5pt]
    \end{tabular}
    }
    \vspace{-2mm}
\end{table}

\noindent
\pamirevises{\textbf{Backbone Flexibility.} We conduct extensive evaluation of various SCD backbone designs to examine the flexibility of \name, including LSTM, 1D convolution, and 1D convolution with residual connections, denoted as \name-lstm, \name-conv, and \name-r-conv, respectively.
For a fair comparison, all variants share the same base structure as \name's DNNs implementation.
Results in Table~\ref{tab:ablation_instance-specific} show that all three enhanced SCD backbones achieve consistently better detection performance, yielding AUCROC improvements of 4.4\%, 3.1\%, and 3.8\%, respectively, compared to the original SCD based on canonical DNNs.
These results confirm that \name's effectiveness is not tied to a specific backbone architecture, demonstrating its architecture-agnostic design and flexibility for further performance improvements.}

\noindent
\textbf{Focal objective function.}
To assess the effectiveness of the focal objective function $\mathcal{L}_{IEC}$, we introduce \name-ce, a variant trained with cross-entropy loss for the IEC module. 
\pamirevises{Comparing \name-ce with \name reveals that incorporating the focal term in the loss function yields an average 2.7\% performance improvement across six datasets in Table~\ref{tab:ablation_instance-specific}.}
Given the inherent class imbalance in anomaly detection tasks and the imbalance in our pseudo labels, the focal term enhances IEC training, leading to superior performance.

\noindent
\textbf{Prior knowledge.}
We introduce a variant \name-pl, which enhances the pseudo-labeling strategy by incorporating 1\% of labeled anomalies into the training set to evaluate \name's adaptability and generalization in situations with limited labeled samples.
The results in Table~\ref{tab:ablation_instance-specific} reveal that the evidential IEC successfully leverages prior knowledge and considerably enhances the learning capacity of \name by incorporating only a small number of labeled samples.

\begin{table}[t!]
    \centering
    \renewcommand{\arraystretch}{1.2}
    \caption{ \pamirevises{Performance on dedicated ablation studies measured by AUCROC.}}
    \vspace{-2mm} \label{tab:ablation_instance-specific}
    \resizebox{1.0\columnwidth}{!}{
        \begin{tabular}{ l ccc c c c }
    \toprule[1.5pt]
   	\multirow{2}{*}{Variant} & \multirow{2}{*}{Ion.} & \multirow{2}{*}{BGL} & \multirow{2}{*}{M.T.} &  INSECTS  &   INSECTS  & INSECTS \\    
    & & & & -Mix & -Abr& -Inc \\
		\hline
    \name-lstm & 0.976 & 0.925 & 0.912 & 0.815 &0.918 &0.859 \\
 \name-conv & 0.982 & 0.924 & 0.893 & 0.811 & 0.889 & 0.838 \\
  \name-r-conv & 0.986 & 0.932 & 0.892 & 0.807 & 0.916 & 0.841 \\
    \hline
    \name-ce & 0.961 & 0.898 & 0.852 & 0.739 & 0.802 & 0.787 \\
    \name-pl & 0.982 & 0.932 & 0.872 & 0.802 & 0.842 &  0.831 \\
    \name-re & 0.931 & 0.816 & 0.725 & 0.628 & 0.682 & 0.536 \\
    \hline
    \name-$\mathcal{R}$ & 0.948 & 0.860 & 0.842 & 0.755 & 0.817 &  0.798 \\
    \name-$\mathcal{Q}$ & 0.962 & 0.864 & 0.846 & 0.761 & 0.829 & 0.806 \\
    \name-$\mathcal{A}$ & 0.967 & 0.877 & 0.871 & 0.769 & 0.818 & 0.810 \\
    \hline
    \name & 0.972 & 0.906 & 0.866 & 0.785 & 0.832 & 0.814 \\
    \bottomrule[1.5pt]
    \end{tabular}
    }
\end{table}

\begin{table}[t!]
   \small
    \centering
    \renewcommand{\arraystretch}{0.8}
    \caption{\blue{Training and inference efficiency of \name.}}
    \vspace{-2mm}
    \label{tab:efficiency}
    
    \resizebox{1.0\columnwidth}{!}{
        \begin{tabular}{c c c c c c}
    \toprule[1.5pt]
    \multirow{2}{*}{Dataset} & \multicolumn{2}{c}{Throughput} & Training & \multirow{2}{*}{Memory (MiB)} \\
   &  Training  &  Inference & Time (s) &    \\
    
    \midrule[0.5pt]
    Ion. &  8,962 &  165,120  & 0.009 & 6.20 \\
    NSL &  140,662   &  77,893,957  &0.022 & 16.28\\
    M.T. &  300,918 &  2,068,214 & 0.018 & 9.71\\
    CPU &  205,884  & 4,300,618 & 0.016 & 9.22 \\
    INSECTS-Abr & 520,218 & 21,576,902 & 0.017 & 38.08\\
    \bottomrule[1.5pt]
        \end{tabular}
    }
    \vspace{-3mm}
\end{table}

\noindent
\textbf{Instance-specific information.}
\pamirevises{A dedicated ablation study is conducted to evaluate the superiority of the instance-aware hypernetwork in DSD, where parameter generation is explicitly conditioned on the current input instance, as opposed to using conventional random embeddings (denoted as \name-re).
As shown in Table~\ref{tab:ablation_instance-specific}, incorporating instance-specific information significantly improves \name's detection performance.
In contrast to the original hypernetwork design, which relies on random embeddings and thus exhibits a weak correlation between parameter generation and the current input, DSD takes into account instance-specific information, enabling more precise and adaptive modeling for online anomaly detection under concept drift.
}

\noindent
\textbf{Dynamic thresholding.}
\pamirevises{To analyze the mechanism design underlying the proposed dynamic thresholding in DTO, we introduce three refined variants with the results summarized in Table~\ref{tab:ablation_instance-specific}.
\name-$\mathcal{R}$ relies solely on reconstruction error for anomaly scoring, resulting in notable performance deterioration.}
This highlights the crucial role of concept uncertainty in refining anomaly assessment by calibrating the anomaly degree based on reconstruction error.
\pamirevises{\name-$\mathcal{Q}$ adopts quantile-based dynamic thresholding $\mu^0_a$, guaranteeing an upper bound on the false positive rate.
However, it becomes overly conservative under high uncertainty, resulting in degraded overall performance due to increased false negatives, which underscores the necessity of the uncertainty-aware regularization term in DTO.
\name-$\mathcal{A}$ replaces the median anomaly score of the sliding window $\mathcal{W}_C$ with the mean to calculate the regularization term $\mu^r_a$.
Although it achieves average performance comparable to \name, it degrades on five out of six datasets, indicating that the median-based design is more robust for threshold regularization under concept drift.}

\subsection{Efficiency}
\label{sec:exp_efficiency}
We evaluate \name's training and inference efficiency by quantifying its throughput, defined as the number of samples processed per second, across benchmark datasets.
As shown in Table \ref{tab:efficiency}, \name maintains high efficiency in both training and inference phases across datasets of varying sizes and dimensions, demonstrating its computational effectiveness and supporting rapid response for OAD applications.

Another important consideration is the training time and the peak memory usage.
Table \ref{tab:efficiency} reports the average training time for each epoch and the peak memory. 
The results demonstrate that \name requires negligible training time and takes low memory usage across these datasets.
This is mainly due to the lightweight design of the key modules of \name, as discussed in detail in Section~\ref{sec:analysis_and_discussion}.
To provide further insights into this matter, we conduct tests to assess the efficiency impact of the IEC and DSD modules. Specifically, we compare peak memory usage and average training time per epoch on CPU of \name with and without IEC and DSD.
The results indicate that the introduction of the IEC and DSD modules leads to a negligible increase in training time by 0.014s and 0.007s, respectively, and peak memory usage by 5.17MB and 2.61MB, respectively.

\begin{figure}[t]
    \centering
\includegraphics[width=1\linewidth]{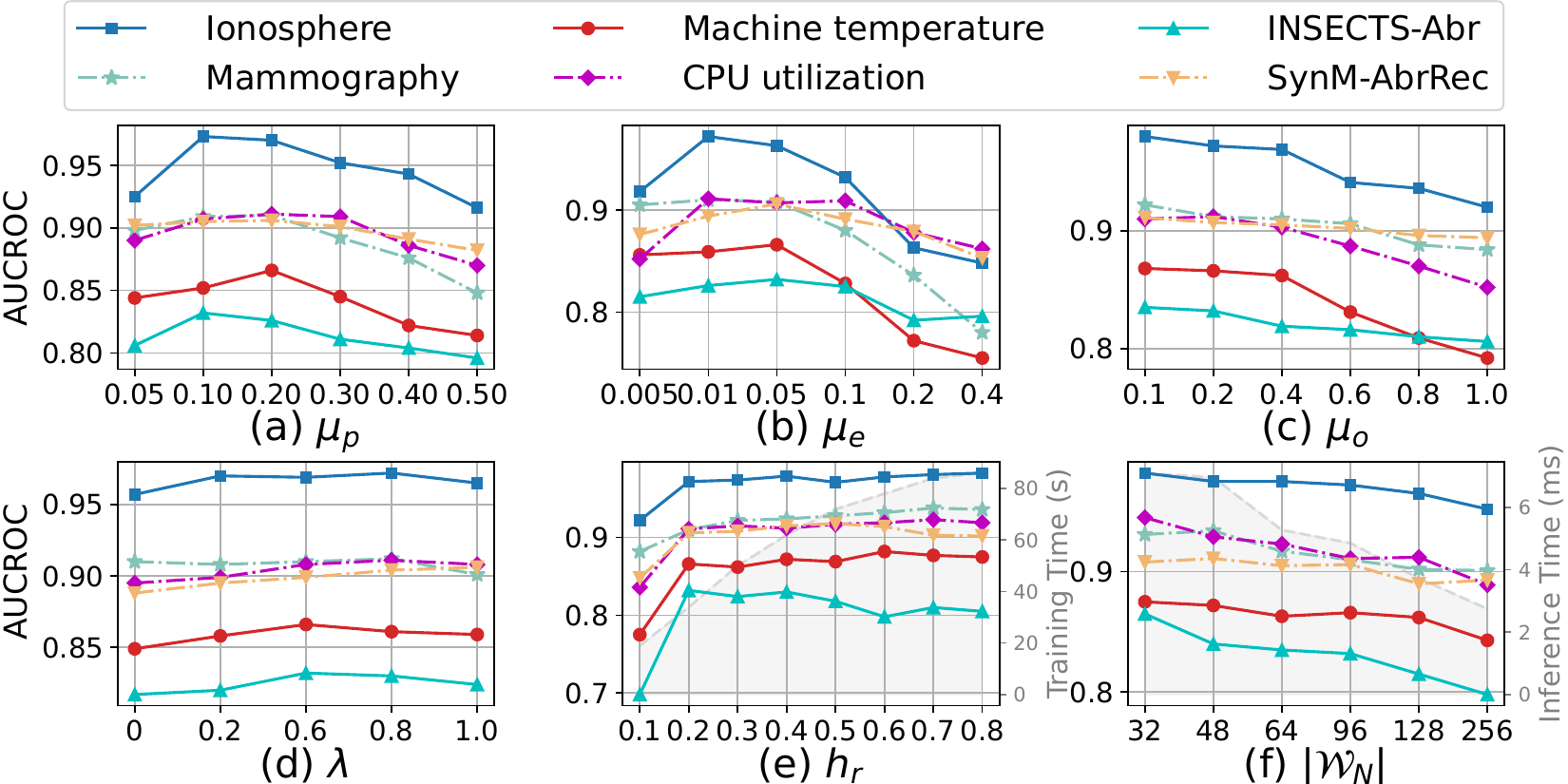}
\caption{\pamirevise{Sensitivity test of key hyperparameters.
    }}
    \label{fig:sentitive}
 \vspace{-3mm}
\end{figure}

\pamirevise{
\subsection{Sensitivity Study}
\label{sec:exp_sensitivity}
We first conduct a sensitivity study on three threshold hyperparameters $\mu_{p}$, $\mu_{e}$, and $\mu_{o}$.
As shown in Figure~\ref{fig:sentitive}, the performance curves of the pseudo-labeling threshold $\mu_{p}$ and uncertainty threshold $\mu_{e}$ across six datasets exhibit an initial increase followed by a gradual decline. This pattern arises because setting these thresholds too low assigns an excessive number of samples to the DSD module even under stable concepts, whereas overly high thresholds delay or suppress timely adaptation to concept drift.
The parameter $\mu_{o}$, representing the ratio of samples exceeding the update threshold within the sliding window $\mathcal{W}_N$, shows mild fluctuations but remains generally stable. Smaller $\mu_{o}$ values improve detection accuracy but may trigger frequent offline updates, thereby reducing efficiency.
In practice, when prior data characteristics are unknown, we recommend setting $\mu_{p}$ between 0.1 and 0.2, $\mu_{e}$ within 0.01–0.05, and $\mu_{o}$ between 0.2 and 0.4 to balance detection performance and computational cost.}

\pamirevise{Additionally, we analyze the sensitivity of $\lambda$, which controls the strength of uncertainty calibration in the anomaly score. The results show mild sensitivity across all datasets, with performance rising initially and slightly declining afterward, and $\lambda$ values between 0.6 and 0.8 yielding the most consistent results.
We further evaluate the historical data ratio $h_r$ within the range of 0.1 to 0.8 and observe that low ratios lead to notable performance degradation due to insufficient information for learning central concepts. Performance improves as $h_r$ increases, whereas overly large ratios bias the model toward outdated concepts, reducing adaptability to new concepts and increasing training cost. To balance effectiveness and efficiency, we set $h_r=0.2$ in this work.
Similarly, the window size $|\mathcal{W}_N|$ critically balances accuracy and efficiency in DTO. As shown in Figure~\ref{fig:sentitive}, smaller windows ensure timely updates but increase inference time, while larger ones reduce cost at the expense of slower adaptation. Overall, \name maintains stable performance across a broad range of $|\mathcal{W}_N|$, enabling flexible adjustment for different practical needs.
}

\begin{figure}[t]
    \centering
    \includegraphics[width=1\linewidth]{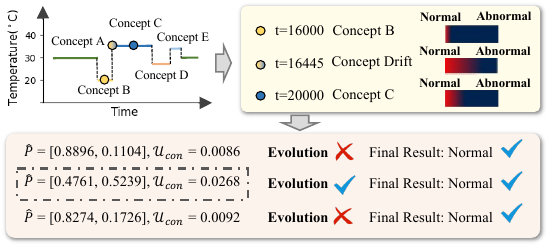} 
\caption{\pamirevise{\scalebox{0.93}[1]{Interpretation of model behavior across evolving concepts.}}
    }
    \label{fig:interpretability}
    \vspace{-4mm}
\end{figure}

\pamirevise{
\subsection{Interpretability}
\label{sec:exp_interpretability}
To provide a semantic interpretation, Figure~\ref{fig:interpretability} illustrates \name's behavior across evolving concepts, showing how probability shifts and uncertainty dynamics drive adaptation.
At t = 16000, \name exhibits low entropy and uncertainty, correctly recognizing the sample as normal.
At t = 16445, high concept uncertainty indicates a distributional shift, activating the IEC’s dynamic mode.
At t = 16445, \name yields a probability distribution marked by significant concept uncertainty, suggesting a potential for concept drift that warrants attention.
In response, \name transitions to the dynamic mode and adjusts base parameters in real time, accurately classifying the sample as normal.
By t=20000, we can observe that 
the model has effectively adapted to the new concept, as evidenced by low uncertainty in its predictions.
These results show that \name achieves interpretable decisions while decoupling drift detection from anomaly identification and tracking state changes via uncertainty estimation.}

\begin{figure}[t]
    \centering
     \includegraphics[width=1\linewidth]{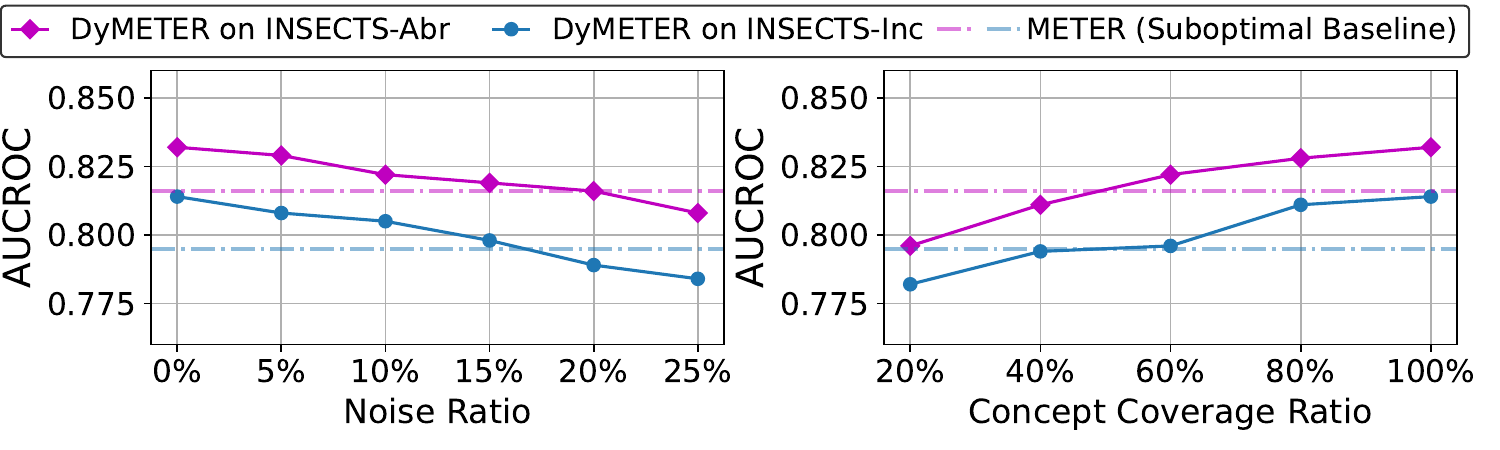} 
    \vspace{-5mm}
\caption{\pamirevise{\scalebox{0.95}[1]{Robustness under noisy and insufficient concept coverage.}}
    }
    \label{fig:robust}
    \vspace{-4mm}
\end{figure}

\pamirevise{
\subsection{Robustness}
\label{sec:exp_robustness}
To further examine the practical robustness of \name, we simulate two challenging yet realistic scenarios:
(i) Noisy initialization, where we inject Gaussian noise into training while keeping the training set size unchanged, mimicking environments with unstable or mislabeled records, and
(ii) Insufficient concept coverage, where only a subset of the original training set is used, limiting the representation of the nominal concept.
The results summarized in Figure~\ref{fig:robust} span representative drift settings, abrupt and incremental drift, evaluated on the INSECTS-Abr and INSECTS-Inc, respectively.
Although performance slightly declines, the degradation remains modest and smooth, and \name still consistently outperforms the suboptimal baseline (METER) over a wide range of noise and coverage levels, confirming robustness even under imperfect initial concept learning.
The robustness of \name stems from its hierarchical and uncertainty-aware training design, where IEC is trained only on low-uncertainty samples to prevent noise propagation as Eq.~(\ref{eq:pseudo_label}), while DTO continuously recalibrates decision boundaries to absorb score shifts and sustain stability, with offline updates providing a contingency mechanism under extreme conditions.
}


\begin{figure}[t]
    \centering
     \includegraphics[width=1\linewidth]{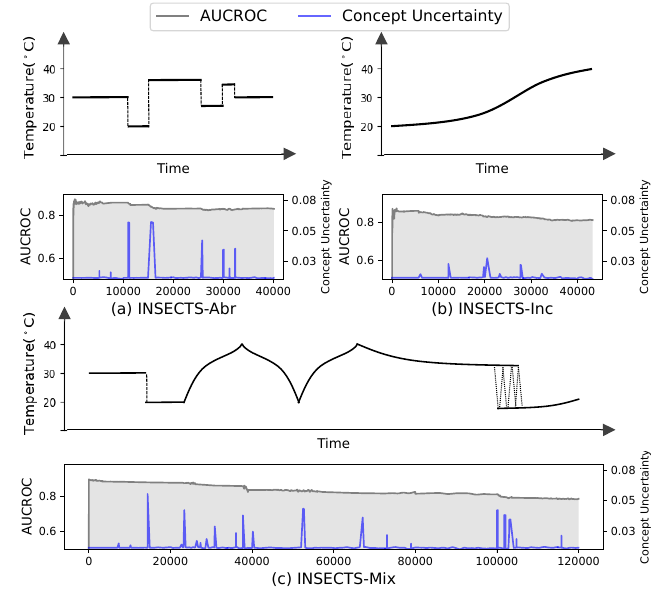} 
    \vspace{-5mm}
\caption{\pamirevise{In-depth analysis on concept drift adaptation.}
    }
    \label{fig:concept_drift}
    \vspace{-4mm}
\end{figure}

\pamirevise{
\subsection{In-depth Analysis on Concept Drift Adaptation}
\label{sec:concept_drift}
To gain deeper insight into \name's behavior in evolving environments, we perform an in-depth analysis of its dynamic performance across different drift scenarios.
First, we visualize the temporal evolution of AUCROC and concept uncertainty along the data stream under abrupt and incremental drift settings, where changes in temperature serve as drift indicators following~\cite{souza2020challenges}.
As illustrated in Figure~\ref{fig:concept_drift}, \name consistently exhibits high AUCROC without notable degradation throughout the data stream.
Although the uncertainty curve exhibits mild fluctuations, it rises sharply at drift onset with virtually no delay, validating the model's responsiveness to evolving concepts.
To further assess adaptability under more complex conditions, we construct a large-scale dataset with mixed drift types by sequentially concatenating four INSECTS variants.
This setup produces a unified data stream where multiple drift types (i.e., abrupt, incremental, gradual, and reoccurring) co-exist.
To ensure label-space consistency and smooth transitions between adjacent data streams, mixed sampling is applied at their junctions.
As shown in Figure~\ref{fig:concept_drift}(c), although the AUCROC exhibits slight declines over time (e.g., around t=40000 and t=100000), the sharp rise in uncertainty at drift onsets promptly activates DSD for online adaptation, enabling \name to sustain competitive overall performance (AUCROC above 0.785) across complex drift dynamics.
}

\subsection{\name in Action}
\label{sec:flink}
\begin{figure}[t]
    \centering
    \vspace{-2mm}
    \includegraphics[width=1\linewidth]{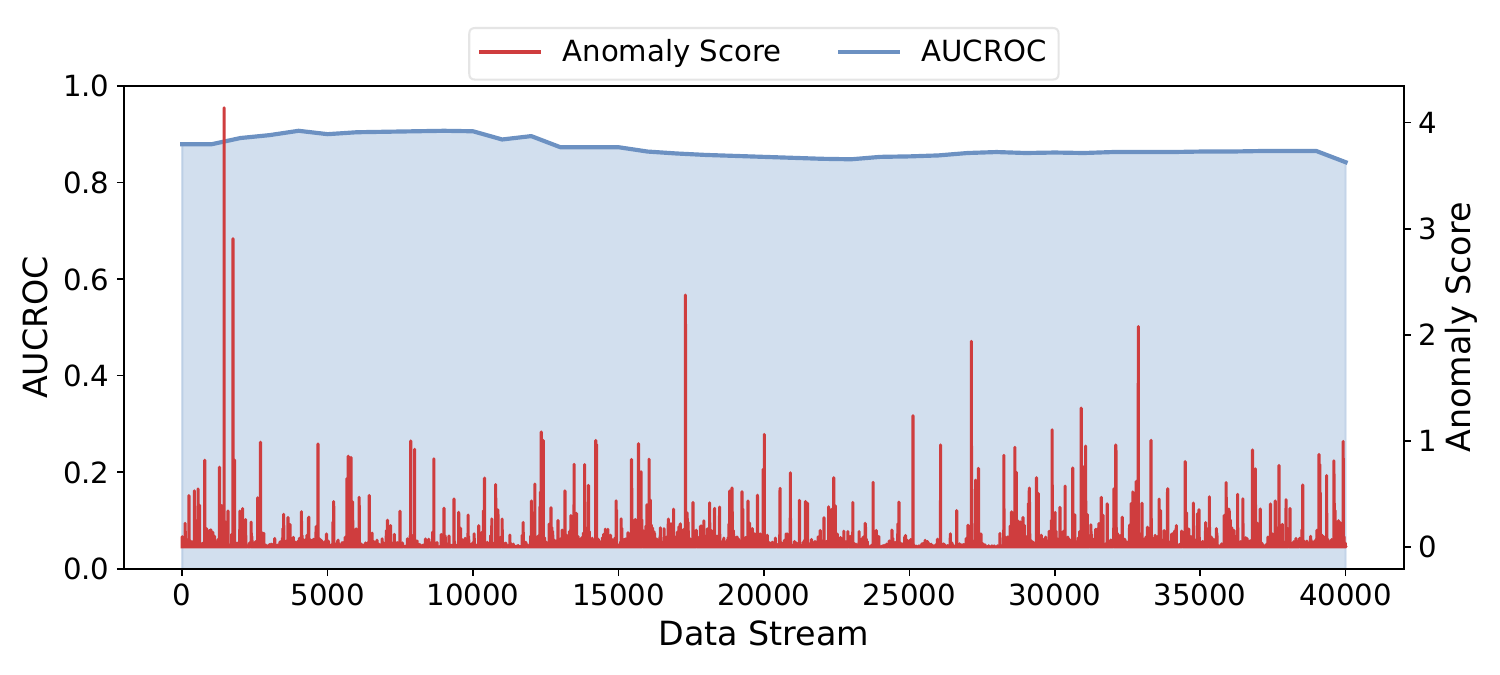}
    \vspace{-4mm}
    \caption{\blue{Timely performance of \name on Flink.} }
    \label{fig:flink}
    \vspace{-5mm}
\end{figure}

To demonstrate the practical application of \name, we integrated it into Apache Flink, a sophisticated framework for stateful computations across both unbounded and bounded data streams~\cite{carbone2015apache}. 
\name operates within a dedicated Flink operator that not only manages the necessary computational environment and resources but also facilitates data interactions through versatile connectors and ensures robust failure management.
This integration enables \name to support real-time data processing pipelines, perform large-scale exploratory data analysis, and execute efficient ETL processes.
Our experiments on the INSECT-Abr dataset showcase \name's seamless integration within Flink in a live setting, as shown in Figure~\ref{fig:flink}.
Notably, \name consistently achieves a high AUCROC level, showcasing its reliability for ongoing and effective anomaly detection in real-time operational environments.

\ignore{
Stream processing engines play a pivotal role in deploying
OAD frameworks.
To illustrate how \name can function as a component of a larger-scale system, we integrate \name into Apache Flink~\cite{carbone2015apache}, a framework and distributed processing engine for stateful computations over unbounded and bounded data streams. 
Flink wraps \name inside a Flink operator, where Flink helps to establish the required environment, manage resources, read/write the data with versatile connectors, and handle failures.
Building streaming workloads upon Flink empowers \name with real-time data processing pipelines, large-scale exploratory data analysis, and ETL processes.
The experiments are conducted on the INSECT-Abr dataset.
The timely output results, as illustrated in Figure~\ref{fig:flink}, validate a seamless integration of \name with a stream processing engine, confirming its efficacy in supporting real-time anomaly detection.
}

\section{Related Work}
\label{sec:related work}

\noindent
\textbf{Anomaly Detection.}
Anomaly detection (AD) has been extensively studied in various fields such as intrusion detection \cite{sood2023intrusion,zhu2022adaptive,zhu2024llms}, healthcare \cite{su2021few,vsabic2021healthcare, zhu2023uaed}, and finance \cite{xiao2024vecaug,hilal2022financial}.
\pamirevise{With the rise of deep learning, reconstruction-based approaches leveraging autoencoders~\cite{vu2019anomaly,zhu2022adaptive}, variational autoencoders (VAEs)~\cite{li2020anomaly,zhu2023uaed}, and generative adversarial networks (GANs)~\cite{zhang2023stad} have become dominant, detecting anomalies as instances with large reconstruction errors.
More recently, diffusion-based models~\cite{chen2023imdiffusion,he2024diffusion,zhang2023unsupervised} have emerged, further advancing AD through superior reconstruction quality.
However, their reliance on extensive training data and inability to adapt to evolving data streams make them unsuitable for online anomaly detection.
}

\pamirevise{
\noindent\textbf{Online Anomaly Detection.}
Online anomaly detection (OAD) aims to identify abnormal behavior in evolving data streams.
Unlike traditional AD, OAD faces the challenge of concept drift.
Early studies~\cite{bifet2007learning,cavalcante2016fedd,huang2014detecting} primarily rely on sliding windows to monitor statistical shifts (e.g., mean or variance) in streaming features.
However, such methods are insufficient for addressing the more intricate distributional shifts, correlations, and pattern transformations that characterize real-world drifts.
To handle evolving streams, various AD paradigms have been extended for online use, giving rise to two dominant OAD categories: incremental and ensemble methods.}

\pamirevise{
Incremental methods~\cite{guha2016robust, na2018dilof, bhatia2021mstream, bhatia2022memstream,boniol2021sand} construct an initial detector and progressively update it as new data arrives.
Several algorithms rely on similarity measures to detect anomalies in evolving sequences~\cite{guha2016robust,boniol2021sand}, while deep learning models extend this paradigm by capturing non-linear feature dependencies for richer representations.
For instance, MStream~\cite{bhatia2021mstream} and its deep variant MStream-AE employ autoencoder-based representations to improve detection accuracy, while MemStream~\cite{bhatia2022memstream} integrates a denoising autoencoder with a memory module to model evolving data distributions.
However, incremental methods remain computationally inefficient and prone to adaptation delays, as they require frequent model updates through retraining or fine-tuning to accommodate new concepts.
}

\pamirevise{
Ensemble methods~\cite{tan2011fast, ding2013anomaly, sathe2016subspace, pevny2016loda,manzoor2018xstream,yoon2022adaptive,mirsky2018kitsune,gopalan2019pidforest} maintain multiple models specialized for different concepts and aggregate their outputs for robust detection under evolving distributions.
iForestASD~\cite{ding2013anomaly} combines sliding windows with isolation forests to detect anomalies in streaming data, while xStream~\cite{manzoor2018xstream} employs an ensemble of randomized projections that are continuously updated to accommodate temporal shifts.
Beyond these methods, deep learning–driven ensembles have further advanced this paradigm.
For instance, Kitsune~\cite{mirsky2018kitsune} employs multiple autoencoders to collaboratively detect anomalies in network streams, and ARCUS~\cite{yoon2022adaptive} introduces adaptive model pooling to manage multiple deep models for time-varying concept drifts.
However, the effectiveness of ensemble methods is constrained by the number and diversity of constituent models, whose maintenance incurs considerable computational and memory overhead.}


\pamirevise{
Beyond these paradigms, recent studies have proposed drift-adaptive architectures that embed adaptation mechanisms directly into model design~\cite{wang2023drift,dai2024sarad,li2024state,kim2024model,nie2024dynamic}.
For instance, D$^3$R~\cite{wang2023drift} addresses drift through dynamic decomposition and diffusion-based reconstruction, while SARAD~\cite{dai2024sarad} employs Transformer-based spatial–temporal association modeling to capture evolving inter-feature relationships.}
\pamirevises{While these approaches alleviate the reliance on repeated tuning and model ensembles, they still lack online fine-grained adaptability and robust threshold calibration under evolving concepts.
In contrast, \name advances this line of research by integrating instance-aware inference-time adaptation with dynamic threshold optimization, enabling effective, efficient, and interpretable anomaly detection across diverse drift scenarios.}
\section{\pamirevise{Limitations and Future Directions}}
\label{sec:limitations}

\pamirevise{
While \name demonstrates strong adaptability and robustness across diverse drift scenarios, several limitations remain that open avenues for future research.
The current framework assumes that historical data provide a reasonably representative basis for the underlying nominal concept. In highly noisy or data-scarce environments, this assumption may not always hold, potentially leading to an imperfect initialization.
Nevertheless, the robustness experiments in Section IV-H show that \name remains resilient under such adverse conditions, consistently achieving competitive performance against the suboptimal baseline.
Future extensions could incorporate self-supervised or regularization strategies to further enhance initialization robustness and reduce reliance on clean historical data.
Another promising direction is to advance \name toward proactive adaptation, where predictive uncertainty dynamics are leveraged to anticipate concept drift events before they occur, ultimately enabling lifelong anomaly detection in evolving real-world environments.
}
\section{Conclusions}
\label{sec:conclusion}
In this paper, we propose a novel online anomaly detection (OAD) framework \name that effectively, efficiently, and interpretably addresses the concept drift challenge in evolving data streams.
\pamirevise{\name integrates a static detector, trained on historical data, to capture recurring central concepts, while dynamically adapting to emerging ones through a hypernetwork-based dynamic detector.
To achieve efficient and interpretable model evolution, an intelligent evolution controller is introduced to estimate concept uncertainty and guide adaptive updates.
Furthermore, a dynamic threshold optimization strategy enables adaptive decision boundaries that continuously adjust to emerging concepts in real time.}
Extensive experimental evaluations demonstrate that \name consistently outperforms state-of-the-art OAD methods across diverse scenarios while facilitating valuable interpretability.

\ignore{
In this paper, we present a novel framework \name for online anomaly detection (OAD), which addresses the challenge of concept drift in data streams in an effective, efficient, and interpretable manner.
By leveraging a static concept-aware detector trained on historical data, \name captures and handles recurring central concepts, while dynamically adapting to new concepts in evolving data streams using a lightweight drift detection controller and a hypernetwork-based parameter shift technique. 
The evidential evolution controller enables efficient and interpretable concept drift detection.
The proposed framework is equipped with a dynamic threshold optimization strategy enabling adaptive decision boundaries corresponding to various emerging concepts.
Our experimental study demonstrates that \name consistently outperforms existing OAD approaches in various scenarios and facilitates valuable interpretability. }




\bibliographystyle{IEEEtran}
\bibliography{reference}

\vspace{-10mm}
\begin{IEEEbiography}[{\includegraphics[width=1in,height=1.25in,clip,keepaspectratio]{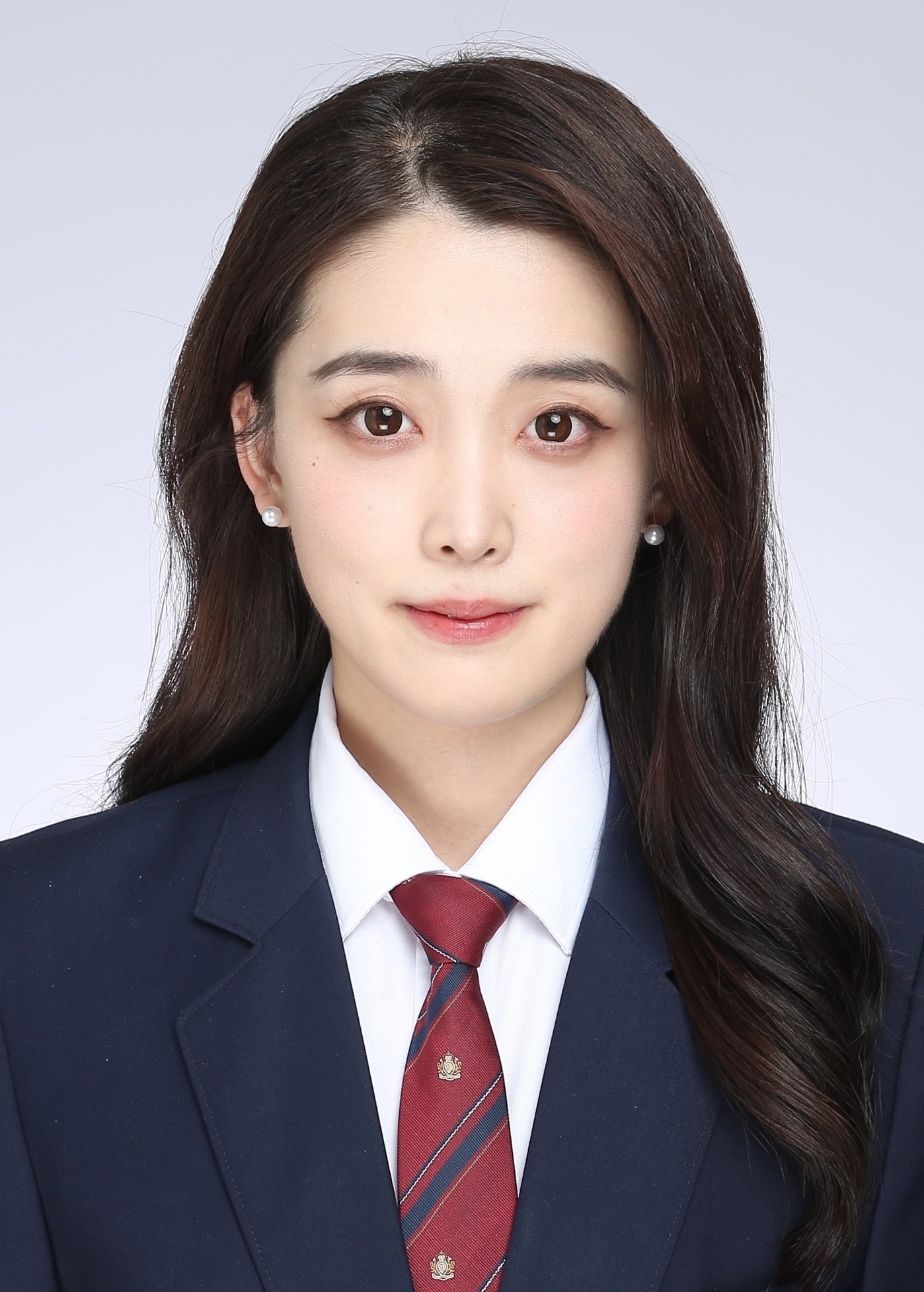}}]{Jiaqi Zhu}
received the B.E. degree in automation from the Nanjing Institute of Technology, Nanjing, China, in 2019. She is currently a Ph.D. student of School of Automation at the Beijing Institute of Technology, Beijing, China.
Her research interests include deep learning, anomaly detection, and time series analysis.
\end{IEEEbiography}

\vspace{-10mm}

\begin{IEEEbiography}[{\includegraphics[width=1in,height=1.25in,clip,keepaspectratio]{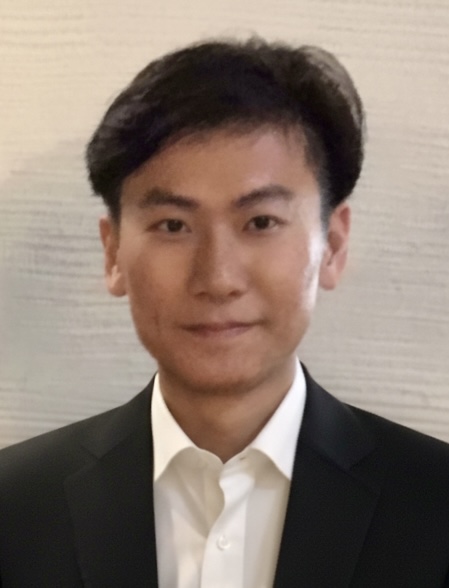}}]{Shaofeng Cai}
received the B.Sc. degree in computer science from Peking University, China, and the Ph.D. degree in computer science from the National University of Singapore, in 2016 and 2021, respectively.
He is currently a senior research fellow with the School of Computing, National University of Singapore.
His research interests include adaptive neural architectures and techniques, in-database analytics, tabular deep learning, and anomaly detection.
\end{IEEEbiography}

\vspace{-10mm}
\begin{IEEEbiography}[{\includegraphics[width=1in,height=1.25in,clip,keepaspectratio]{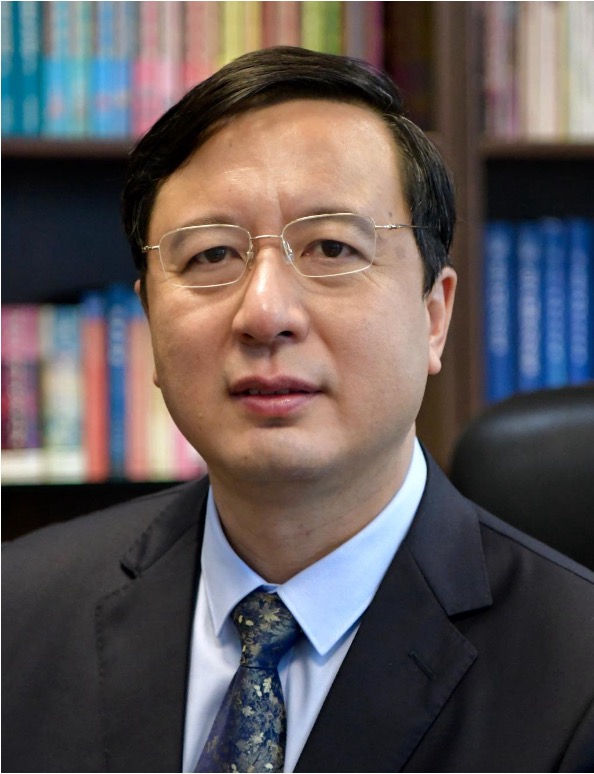}}]{Jie Chen}
(Fellow, IEEE) received the B.Sc., M.Sc., and Ph.D. degrees in Control Theory and Control Engineering from the Beijing Institute of Technology, Beijing, China, in 1986, 1996, and 2001, respectively. 
He was the President of Tongji University, Shanghai, China, during 2018-2023.

He is currently a Professor with the Harbin Institute of Technology, Harbin, China, and serves as the Director of the National Key Lab of Autonomous Intelligent Unmanned Systems at the Beijing Institute of Technology.
His research interests include complex systems,
multiagent systems, multiobjective optimization
and decision, constrained nonlinear control, and optimization methods.
He is currently the Editor-in-Chief of Unmanned Systems and the Journal of Systems Science and Complexity. He has served on the editorial boards for several journals, including the IEEE Transactions on Cybernetics, International Journal of Robust and Nonlinear Control, and Science China Information Sciences. 

He is a Fellow of IFAC, and a member of the Chinese Academy of Engineering.
\end{IEEEbiography}

\vspace{-10mm}

\begin{IEEEbiography}[{\includegraphics[width=1in,height=1.25in,clip,keepaspectratio]{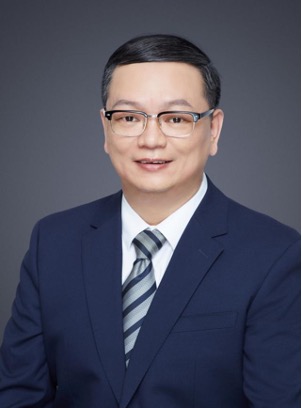}}]{Fang Deng}
(Fellow, IEEE) received the B.E. and Ph.D. degrees in control science and engineering from the Beijing Institute of Technology, Beijing, China, in 2004 and 2009, respectively.

He is currently a Professor at the School
of Automation, Beijing Institute of Technology.
His current research interests include machine learning, intelligent information processing, and control of renewable energy resources.
\end{IEEEbiography}

\vspace{-10mm}

\begin{IEEEbiography}[{\includegraphics[width=1in,height=1.25in,clip,keepaspectratio]{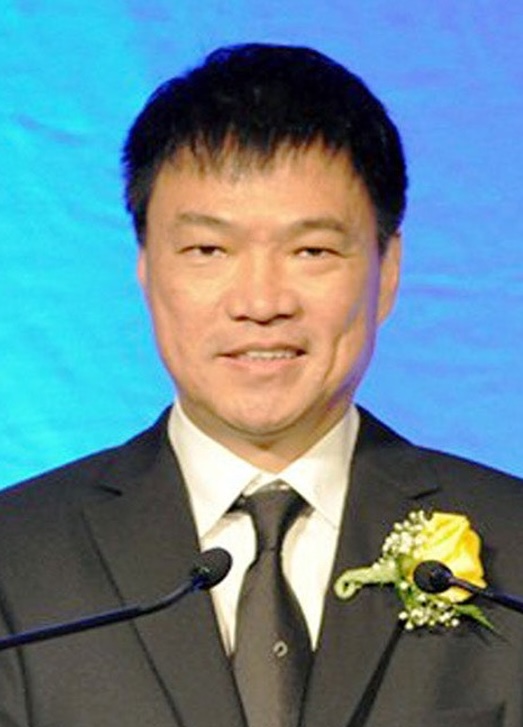}}]{Beng Chin Ooi}
(Fellow, IEEE) received the B.Sc. and Ph.D. degrees from Monash University, Australia, in 1985 and 1989 respectively.
He is the Lee Kong Chian Centennial professor and an NGS faculty member with the National University of Singapore.
His research interests include database system architectures, performance issues, indexing techniques, and query processing, in the context of multimedia, spatio-temporal, distributed, parallel, blockchain, and in-memory systems. 
He served as the editor-in-chief of the IEEE Transactions on Knowledge and Data Engineering (2009-2012), a trustee board member, and the president of the VLDB Endowment (2014-2017).

He is a fellow of ACM, CCF, Singapore National Academy of Science, and Singapore Academy of Engineering.
He is a foreign member of Academia Europaea and Chinese Academy of Sciences.

\end{IEEEbiography}

\vspace{-10mm}

\begin{IEEEbiography}[{\includegraphics[width=1in,height=1.25in,clip,keepaspectratio]{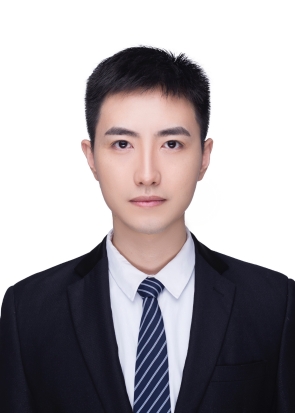}}]{Wenqiao Zhang} received the Ph.D. degree from the Zhejiang University, Hangzhou, China, in 2021. He is an assistant professor at the College of Software, Zhejiang University, China. His current research interests include cross-media analysis and computer-aided healthcare. So far, he has authored more 50 papers in top-tier scientific journal/conference such as the IEEE Transactions on Visualization and Computer Graphics, AAAI, WWW, NeurIPS, CVPR, ACL, ACM-MM, KDD.
\end{IEEEbiography}
\vspace{-10mm}

\end{document}